\documentclass[lettersize,journal]{IEEEtran}
\usepackage{amsmath,amsfonts}
\usepackage{algorithm}
\usepackage{algpseudocode}
\usepackage{array}
\usepackage[caption=false,font=normalsize,labelfont=sf,textfont=sf]{subfig}
\usepackage{textcomp}
\usepackage{float}
\usepackage{stfloats}
\usepackage{url}
\usepackage{verbatim}
\usepackage{graphicx}
\usepackage{cite}
\usepackage{bm}
\usepackage{multirow}
\usepackage{multicol}
\usepackage{footnote}

\begin{document}
\title{Generalized Anthropomorphic Functional Grasping  with Minimal Demonstrations}


\author{Wei Wei, Peng Wang, Sizhe Wang
\thanks{P. Wang is with Institute of Automation and the School of Artificial Intelligence, Chinese Academy of Sciences, Beijing 100190, China, and with the CAS Center for Excellence in Brain Science and Intelligence Technology, Chinese Academy of Sciences, Shanghai 200031, China, and also with the Centre for Artificial Intelligence and Robotics, Hong Kong Institute of Science and Innovation, Chinese Academy of Sciences, Hong Kong 999077, China (email: peng\_wang@ia.ac.cn).}
\thanks{W. Wei and S. Wang are with Institute of Automation and the School of Artificial Intelligence, Chinese Academy of Sciences, Beijing 100190, China.}
}

\maketitle

\begin{abstract}
This article investigates the challenge of achieving functional tool-use grasping with high-DoF anthropomorphic hands, with the aim of enabling anthropomorphic hands to perform tasks that require human-like manipulation and tool-use. However, accomplishing human-like grasping in real robots present many challenges, including obtaining diverse functional grasps for a wide variety of objects, handling generalization ability for kinematically diverse robot hands and precisely completing object shapes from a single-view perception. To tackle these challenges, we propose a six-step grasp synthesis algorithm based on fine-grained contact modeling that generates physically plausible and human-like functional grasps for category-level objects with minimal human demonstrations. With the contact-based optimization and learned dense shape correspondence, the proposed algorithm is adaptable to various objects in same category and a board range of robot hand models. To further demonstrate the robustness of the framework, over 10K functional grasps are synthesized to train our neural network, named DexFG-Net, which generates diverse sets of human-like functional grasps based on the reconstructed object model produced by a shape completion module. The proposed framework is extensively validated in simulation and on a real robot platform.  Simulation experiments demonstrate that our method outperforms baseline methods by a large margin in terms of grasp functionality and success rate. Real robot experiments show that our method achieved an overall success rate of 79\% and 68\% for tool-use grasp on 3-D printed and real test objects, respectively, using a 5-Finger Schunk Hand. The experimental results indicate a step towards human-like grasping with anthropomorphic hands.

\end{abstract}

\begin{IEEEkeywords}
Functional Tool-use Grasping, Anthropomorphic hands, Learn from Demonstration
\end{IEEEkeywords}

\section{Introduction}
\IEEEPARstart{M}{anipulating} and using tools is one of the most important skills that humans have evolved. The study of functional tool-use grasping has been a long-standing topic in neuroscience and psychology\cite{baber2003cognition}. Recently, a number of works in computer vision and robotics have started to focus on this problem\cite{fang2020learning}. Enabling multi-fingered hands to grasp objects like humans is a significant challenge, not only  in terms of mechanical design but also in grasp planning and control.

Although recent advances in neural network-based methods\cite{shang2020deep,MDN,multifingan,hgcnet} have achieved impressive results for multi-fingered hand grasping, there is a lack of large-scale datasets for robotic functional grasping. On the one hand, manual grasp collection in simulators or real robots is time-consuming and expensive. On the other hand, existing multi-fingered grasp planner\cite{graspit,deep_differentiable_grasp_planner} have difficulty in planning human-like tool-use grasps for high-DoF anthropomorphic hands. Meanwhile, we notice that several datasets for human hand-object interaction have been proposed in the past few years\cite{contactdb}. Therefore, a natural idea is to “transfer” human hand grasp to robotic hand grasp, since anthropomorphic hands share high morphological similarity with human hands.

Various solutions have been proposed to solve the problem of human-to-robot hand grasp mapping. Most of these method were based on handcrafted correspondence\cite{meattini2022human}, such as direct joint mapping, fingertip pose retargeting and hand posture recognition-based mapping. However, these methods  did not take object properties (e.g., object shape) into consideration. Some recent works proposed leveraging the contact map as an optimization constraint to mimic human grasp demonstration, while a large number of pre-sampled grasps are required\cite{contactgrasp}. 

We observe that two intuitive properties may lead to similar human grasping patterns across objects in the same category: i) Humans tend to touch the functional area of target objects for tool-use grasping; ii) Objects in the same category have similar geometric structure and prominent functional parts.



\begin{figure}[t]
    \centering
    \includegraphics[width=0.99\linewidth]{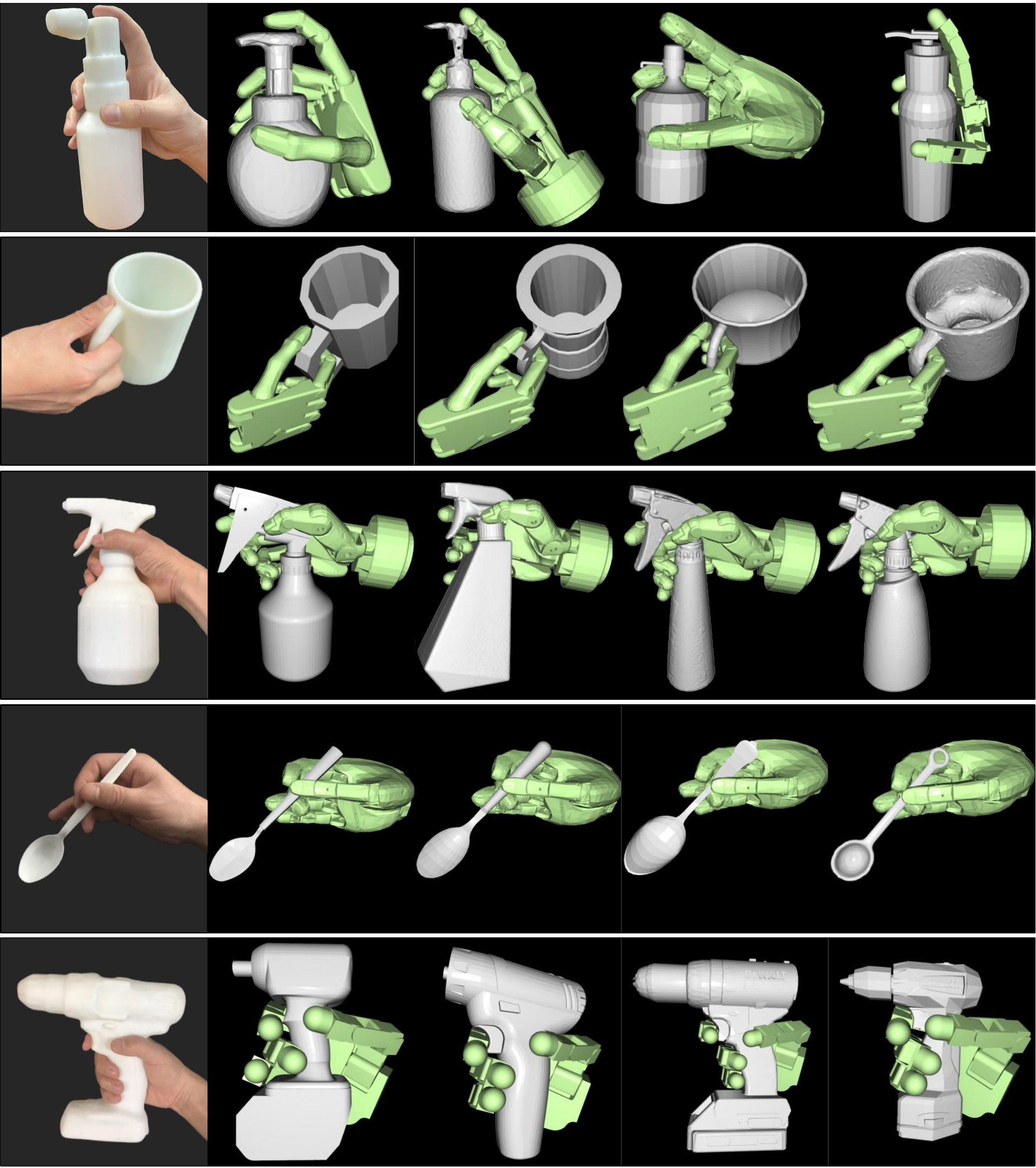}
    \caption{DexFG enables anthropomorphic functional grasping for objects of the same category using a wide range of kinematically diverse anthropomorphic hands, \textit{i.e.} Shadow Hand, Schunk Hand, HIT-DLR Hand and 4-Finger Allegro Hand,  given only one human hand grasp demonstration.}
    \label{fig:idea}
\end{figure}

Inspired by the aforementioned intuitive properties and the high morphological similarity between anthropomorphic hands and human hands, we propose a six-step grasp synthesis pipeline to transfer human grasp demonstrations to multi-fingered hand grasps.  We show that fine-grained contact information extracted from demonstration can be utilized to synthesize human-like tool-use grasps that are adaptable to a wide range of kinematically diverse robotic hands, as shown in Fig. \ref{fig:idea}. Dense shape correspondence across category-level objects is also leveraged to enable contact mapping between intra-category objects, which allows us to synthesize grasps for category-level objects based on only one grasp demonstration. To this end, we build a large-scale synthetic functional grasp dataset, with over 10K grasps, based on the algorithm.


Given the synthetic grasp dataset, we propose a Dexterous Functional Grasp Network (DexFG-Net) for generating human-like functional grasps for tool-use in real-world scenarios. Fig. \ref{fig:framework_overview} presents an overview of the proposed method. The DexFG-Net is capable of generating physically plausible and human-like grasps for target objects based on single-view point cloud input.  Our work stands out from most previous works that focused on planning power grasps for stable hand holding without considering the object's functionality and human grasping habits. Additionally, our work differs from past methods concentrated on fingertip precision grasp with analytical metrics. In summary, this article makes the following contributions:
\begin{itemize}
    \item[1.]A novel functional grasp synthesis algorithm is proposed to produce physically plausible and human-like functional tool-use grasps for kinematically diverse anthropomorphic hands. Furthermore, we build a large-scale functional grasp dataset, with over 10K grasps, for anthropomorphic hand grasping based on the proposed grasp synthesis pipeline. 
    \item[2.]A DexFG-Net is developed to reconstruct the object shape from partial point clouds and generate diverse sets of human-like functional grasps for tool-use. The grasps are further refined to physically plausible grasps for real robot grasp execution. 
    \item[3.]Benchmark experiments conducted in simulation and on real robot platform demonstrate our method show potential to achieve dexterity for human-like grasping with anthropomorphic robotic hands.
\end{itemize}

Benefiting from the generalization ability of the six-step grasp synthesis pipeline and the DexFG-Net, our method could be used for building dataset and planning functional grasps for a wide range of kinematically diverse robot hands.

In this article, we extend our previous work presented in \cite{wei2022dvgg} to a  more challenging task, which involves generating human-like functional grasps for kinematically diverse anthropomorphic hands. Two submodules proposed in \cite{wei2022dvgg} are also used in the DexFG-Net, namely the Variational Grasp Sampler and the Iterative Grasp RefineNet. We made modifications to the input of the RefineNet to fit for this work.  Notably, both of these submodules are trained with novel objective functions that are proposed in this article. 

\section{Related Work}

\subsection{Grasping with Multi-fingered hands}
Existing methods for solving the challenge of multi-fingered hand grasping fall into two categories: analysis-based and learning-based methods.

Analysis-based methods typically generates stable grasps for objects with known shape or approximated primitive shapes. These methods are based on commonly used force-closure metric, $\epsilon$-metric and contact energy optimization\cite{ral2021grasp,kiatos2020geometric}. 
However, due to their restrictive assumptions, such as precise object models and simplified physical conditions, these methods are not applicable for real world scenarios.

Learning-based methods, instead, propose to learn grasping strategies from data. Recent advances in neural network-based methods have achieved impressive results\cite{deep_differentiable_grasp_planner,multifingan} for multi-fingered hand grasping. A typical practice was to sample grasps using extracted primitive shapes of the objects, and preform grasp evaluation based on the local geometry representation of the grasping area\cite{kopicki2019learning,MDN}. In this way, large datasets of object-grasp pairs are required to train the deep neural network. These datasets were generated either manually or by exhaustively searching in simulator with analytic criteria\cite{highdofgrasp}. Another route is based on reinforcement learning\cite{wu2020generative}, which learns the grasping policy in a trial-and-error process but suffering from the gap of sim-to-real migration. 

In contrast to the methods mentioned above, this article mainly focuses on functional grasp synthesis for multi-fingered hands, where the feasible configuration space of functional grasps can be much sparser than that of power grasps.

\subsection{Tool-use Grasping}

Existing works of tool-use grasping can be broadly classified as either task-oriented tool-use grasping or functional object-centric grasping.


 



For task-oriented grasping, previous works \cite{dang2012semantic} proposed to learn task-related semantic constraints that allow for goal-directed grasp selection based on a small dataset of grasp demonstrations for pick and place tasks. Several recent works \cite{ do2018affordancenet} have proposed to train a semantic affordance detection neural network for task-oriented grasping on a large-scale synthetic or manually collected dataset. Furthermore, Fang et al. \cite{fang2020learning} trained a task-oriented grasping network to learn both task-oriented grasping and manipulation policy for the sweeping and hammering tasks, and the learned strategies show robust generalization performance to novel objects.

In contrast to task-oriented grasping, functional tool-use grasping focuses on object-centric functionality, which mainly concerns the functionality and the shape of the target object. Brahmbhatt et al.\cite{contactgrasp} proposed a sample-and-rank grasp synthesis framework, where grasps  sampled by \cite{graspit} are optimized to reproduce consistent contact map demonstrated by human grasp. However, the universal contact map does not specify the unique correspondence between the contact area and the hand segments, which causes the ambiguity in the multiple-to-one mapping. 

In this article, we propose knuckle-level contact association between hand segments and contact area on object surface to avoid the ambiguity caused by  universal contact map representation. Compared to universal contact map used in the above methods, the knuckle-level contact map encodes rich hand-object interaction information, which reflect the contact association between finger segments and the object surface. Moreover, auxiliary anchor points are introduced for precise contact reproduce, which is essential for human-like grasp synthesis with finger side contacts.

\subsection{Human-to-Robot Hand Grasp Mapping}
Modern research has proposed various mechanisms for human-to-robot hand grasp (motion) mapping\cite{meattini2022human}. Hand grasp mapping is not  limited to teleoperation tasks but also used in data collection for high-DoF multi-fingered hands, offering a promising paradigm for robot imitation learning based on human demonstration\cite{dapg}. 

The complete pipeline of hand grasp mapping comprises collecting human hand poses and retargeting them to robotic hand poses. Hand pose capture can be achieved using either vision-based or glove-based solutions.  Vision-based methods involve detecting 2-D or 3-D hand keypoints and then estimating hand poses, whereas glove-based methods use sensor data directly to calculate hand joint angles. 

Robotic hand pose retargeting methods can be categorized into four groups: direct joint mapping, fingertip pose retargeting (cartesian mapping), task-oriented mapping, and hybrid mapping. The direct joint mapping method is primarily utilized for robots with morphological structures similar to that of the human hand, wherein human hand joint angles are assigned to the corresponding joints of the robotic hand. The cartesian mapping mechanism is mainly used for precision grasping with fingertips. Task-oriented mapping methods usually assume a virtual object for contact optimization, but this type of mapping is limited to objects with simple primitive shapes.  Hybrid mapping methods combine the above mapping mechanisms with manually tuned rules for specific situations.

In this research, we introduce a hybrid mapping algorithm designed to translate human hand grasp demonstrations to robotic hands. Our algorithm effectively preserves gesture consistency and accelerates the optimization speed of our grasp synthesis pipeline.

\subsection{Shape Completion for Grasping}
Grasping with single-view sensor input is a challenging task due to incomplete shape perception. As a result, recent efforts have been devoted to addressing the occluded parts of objects through shape completion techniques.

Traditional shape completion methods aim to reconstruct a target object using either geometric symmetry assumptions or heuristic shape approximations\cite{quispe2015exploiting}. The former involves mirroring the object through a symmetry plane. However, the reconstruction results heavily depend on the choice of the symmetry plane and camera perspective, despite the strong symmetry principle of household objects. The latter method reconstructs the shape by fitting primitives to the raw sensor input, typically using a combination of primitives such as boxes and cylinders. However, these methods do not generalize well to complex objects.

Modern methods for shape completion were based on deep neural networks. For instance, Varley et al.\cite{varley2017shape} proposed a large 3-D convolutional neural network to reconstruct the shape from a voxel grid of the point cloud input.  Lundell et al.\cite{lundell2019shape} further proposed a light-weight 3-D CNN architecture based on  Monte-Carlo dropout training strategy. Van der Merwe et al.\cite{van2020learning} proposed a continuous embedding approach using Signed Distance Function (SDF) to jointly predict shape completion and grasp success.

In this research, we train DeepSDF networks\cite{park2019deepsdf} for various object categories, which can encode category-level shape priors, and represent each object shape with an optimized shape vector. Given the network with encoded shape priors, we propose to jointly optimize the object shape and pose using inverse optimization and differentiable SDFRendering technology. Our method significantly outperforms previous methods by leveraging the shape priors constrained optimization.  

\subsection{Dense Visual Correspondence}
Dense visual correspondence plays an vital role in robotic manipulation, including object reconstruction, pose estimation, and part segmentation. Recent studies have revealed that neural networks trained on a large-scale datasets can effectively learn to reason about dense 2-D or 3-D visual correspondence.

Learning 2-D visual correspondence requires image pairs with annotated dense correspondences as training data. A CNN-based network is then trained to learn dense pixel descriptors using discriminative losses. This encourages visual feature embeddings of corresponding pixels to be close and far apart otherwise. However, manual dense correspondence labeling is costly. To address this, many current approaches generate training data through either dense 3D reconstruction\cite{florence2018dense} or simulation synthesis to provide automated correspondence labeling.


For 3-D dense correspondence learning. Chen et al.\cite{chen2019edgenet} proposed to  predict correspondence through point cloud registration, using labeled pairwise correspondence as supervision. To relax constraints on supervision, Bhatnagar et al.\cite{bhatnagar2020combining} proposed to predict part correspondences with part labelling. Furthermore, Cheng et al.\cite{cheng2021learning} proposed to directly learn dense correspondence from the point cloud with self-supervision. 

In this article, an unsupervised approach is utilized to establish dense shape correspondence between intra-category objects by estimating the deformation field of the shape prior. Our method eliminates the need for dense correspondence annotation and can learn in an unsupervised manner.

\section{Problem Statement}
\begin{figure*}[ht]
    \centering
    \includegraphics[width=0.95\linewidth]{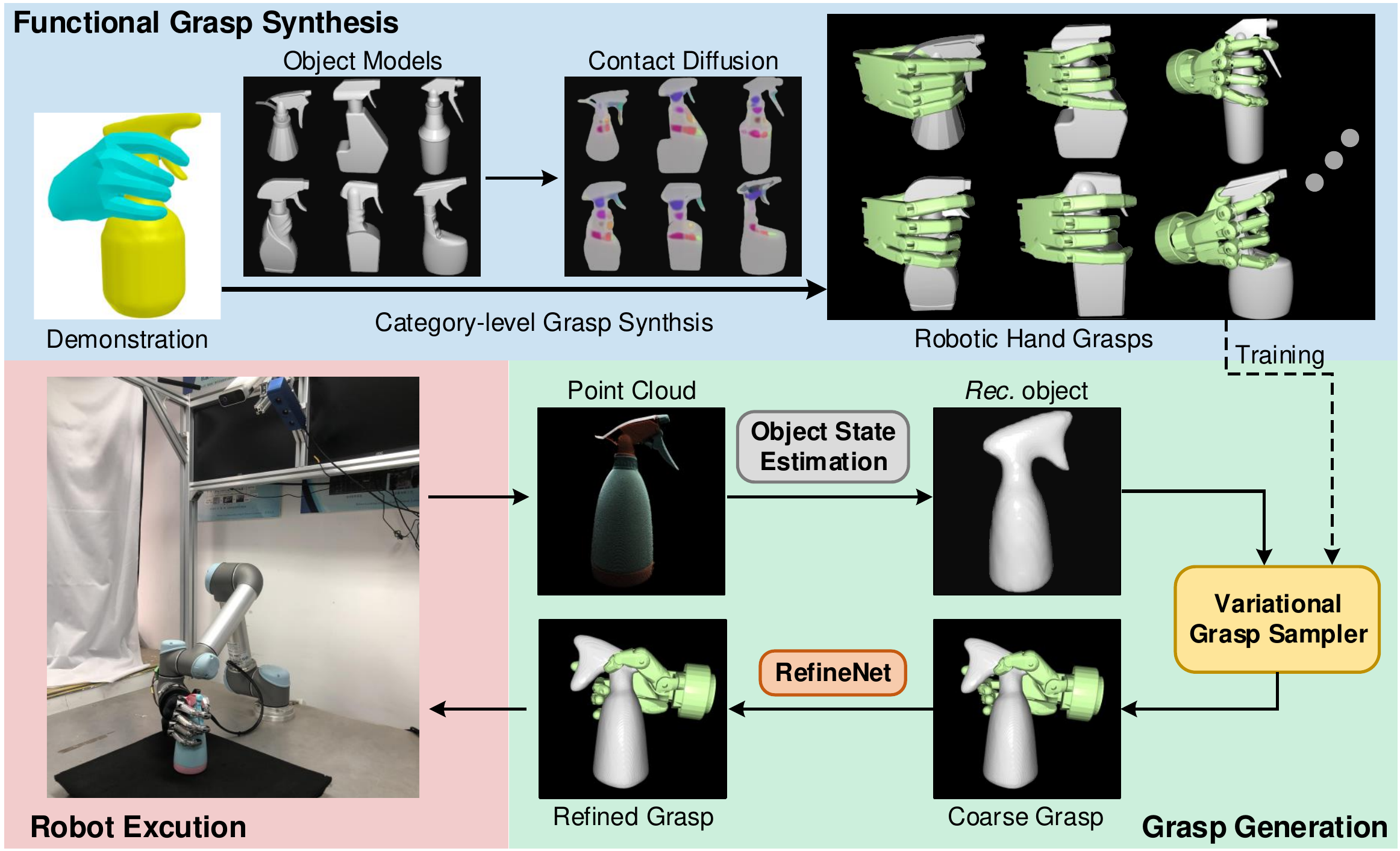}
    \caption{Overview of the proposed DexFG framework for dexterous functional grasping.}
    \label{fig:framework_overview}
\end{figure*}
This article focuses on planning high-DoF functional grasps for tool-use with anthropomorphic hands, which implies generating physically plausible and human-like grasps. As shown in Fig. \ref{fig:framework_overview}, our method takes a 3-D point cloud $\mathcal{P} \in \mathbb{R}^{N \times 3}$ as input and performs a functional grasp to pick up the object. The planning process consists of the following steps: 1) Object reconstruction based on object state parameters estimation; 2) Coarse grasp generation based on a variational grasp sampler; 3) Grasp refinement using an iterative grasp refinement module; and 4) Robust grasp selection for picking up the object. A hand grasp is represented by the preshape $g \in \mathcal{G}$, the preshape is defined as:
\begin{equation}
    \mathcal{G} = \{[q, H], q \in \mathbb{R}^d, H \in SE(3)\}
\end{equation}
where $q$ denotes the hand joint angles, $d$ denotes the hand's degree of freedoms, $H$ represents the hand wrist pose. The composite 3-D hand mesh $\mathcal{M}$ can be obtained by the forward kinematic chain $\mathbb{F}$ as follows:
\begin{equation}
    \mathcal{M} = \mathbb{F}(q, H | [\mathcal{M}^{1}, \mathcal{M}^{2}, \ldots, \mathcal{M}^{i}])
\end{equation}
where $\mathcal{M}^{i}$ denotes the $i$-th link of the hand model.

\subsection{Functional Grasp Synthesis}
The primary task of the functional grasp synthesis algorithm is to synthesize high-DoF human-like grasps for anthropomorphic hands based on human demonstrations. To achieve this, the algorithm initially takes human hand grasp demonstrations as input to extract hand-object contacts.

A human grasp demonstration consists of: 1) A 3-D mesh model of an object $\mathcal{O}$; and 2) A human hand grasp $\mathcal{G}_{h} \in \mathbb{R}^{51}$, which consists of 45 degrees of freedom for hand joint configuration $q_h$, 6 degrees of freedom for hand wrist pose $H_h \in SE(3)$, and its non-rigid deformation mesh model $\mathcal{M}_h$ that is proposed in the MANO hand model\cite{loper2015smpl}. The contact points on the object surface $\mathcal{O}^{C}$ and the human hand surface $\mathcal{M}_h^{C}$ can be derived from the signed distance field between the object and hand mesh.

Given the hand-object contacts and human grasp gesture as reference, the grasp synthesis algorithm optimizes the anthropomorphic hand to have a similar grasp gesture and make similar contacts with the object. We describe the proposed six-step grasp synthesis algorithm in Sec. \ref{sec:functional_grasp_synthesis}.

\subsection{Object Reconstruction}
The objective of object reconstruction is to estimate the shape vector $\bm{c}$ and state parameters $[s, R, T]$ of the target object based on the observed object point cloud $\mathcal{P}_{real}^{\mathcal{O}}$ in a real-world single view perception. The shape vector $\bm{c}$ is used to reconstruct the object mesh with the marching-cube algorithm, based on the shape priors encoded in the DeepSDF network. The state parameters $s, R$, and $T$ represent the scale, rotation, and translation of the object. The shape vector and state parameters are optimized using inverse optimization based on differentiable rendering. A detailed explanation of the process can be found in  Sec. \ref{sec:obj_reconstruction}.

\subsection{Variational Grasp Sampler}
The Variational Grasp Sampler is based on a deep generative model, the Conditional Variational Auto-Encoder (CVAE) \cite{vae}, which has the ability to sample diverse functional grasps $\hat{\bm{G}} =[\hat g_1, \hat g_2, \ldots, \hat g_n]$ given the reconstructed object model. The sampler network is trained on a synthetic functional grasp dataset, which is presented in Sec. \ref{sec:vgs}.

\subsection{Iterative Grasp Refinement}
Given a sampled grasp candidate $\hat g$ and the reconstructed object mesh $\hat{\mathcal{O}}$, the goal of the refinement module is to optimize the grasp $\hat g$ to a grasp $g^{*}$ with less collision and higher functional quality. The network is trained based on the implementation presented in our previous work\cite{wei2022dvgg}, and the objective functions used in this work are  presented in Sec. \ref{sec:igr}


\section{Functional Grasp Synthesis \label{sec:functional_grasp_synthesis}}
Functional grasp synthesis plays a fundamental role in synthesizing our large-scale functional grasp dataset. This article proposes a six-step functional grasp synthesis algorithm to transfer a human grasp demonstration to the grasping of category-level objects using a wide range of anthropomorphic hands.  The algorithm makes the following three assumptions:
\begin{itemize}
    \item[1)]Objects in the same category have similar geometric structure and prominent functional parts.
    \item[2)]All objects are rigid bodies.
    \item[3)]Anthropomorphic hands should touch the functional area for tool-use.
\end{itemize}

The pseudo-code of the six-step synthesis algorithm is depicted in Alg. \ref{alg:cap} and summarized in following order:
\begin{itemize}
    \item[1)]Establish knuckle-level hand-object contact to associate fine-grained contact between hand segments and object surfaces, using both object and hand meshes.
    \item[2)]Define auxiliary anchor points on the hand to align precise finger contacts with target objects.
    \item[3)]Obtain the initial grasp configuration for grasp optimization through human-to-robot hand grasp mapping.
    \item[4)]Obtain dense shape correspondence of category-level objects using a pretrained neural network for contact diffusion between objects of the same category.
    \item[5)]Apply a gradient descent-based algorithm to optimize the initial grasp configuration based on fine-grained hand-object contact objective functions.
    \item[6)]Refine the grasps in simulator to avoid both inter-penetration and self-penetration.
\end{itemize}

Sec. \ref{sec:knuckle_level_contact} through Sec.\ref{sec:functional_hand_grasp} describe the detailed procedures of the functional grasp synthesis algorithm for transferring human-hand grasp demonstrations to a anthropomorphic hand.
\begin{algorithm}
\caption{Functional Grasp Synthesis Algorithm.}\label{alg:cap}
\hspace*{0.15in}{\bf Input:} The human hand demonstration: mesh of the object $\mathcal{O}$, human grasp parameters $g_h$ and human hand mesh $\mathcal{M}_h$, the robot hand model $\mathcal{M}$, the meshes of category-level objects $\{\mathcal{O}_1, \mathcal{O}_2, \cdots, \mathcal{O}_n\}$ and pretrained dense shape correspondence network $\Phi_{\textbf{DSC}}$. \\
\hspace*{0.15in}{\bf Output:} The grasps $\mathcal{G}$ for category-level object models.
\begin{algorithmic}[1]
\For{each $\mathcal{O}_{i} \in \{ \mathcal{O}_{1}, \mathcal{O}_{2}, \cdots \mathcal{O}_{n} \}$}
\State $\Omega_\mathcal{O}$, $\Omega_{\mathcal{M}_h} =$ Get\_contact\_map($\mathcal{O}$, $\mathcal{M}_h$)
\State $\mathcal{O}^{c}$, ${\mathcal{M}_h^c}$ = Get\_contact\_points($\mathcal{O}$, $\mathcal{M}_h$)
\For{$p \in \mathcal{O}^{c}$}
\State $k = \underset{j \in [1, \cdots, N]} {\arg\min} {||p-\mathcal{P}^{\mathcal{M}_h^{j}}||_2^2}$, add $p$ into $\mathcal{O}^{\mathcal{M}_h^k}$
\EndFor
\For{$p \in \mathcal{O}^{c}$}
\State $k = \underset{j \in [1, \cdots, K]} {\arg\min} {||p-{\mathcal{A}_h^{j}}||_2^2}$, add $p$ into $\mathcal{O}^{\mathcal{A}_h^k}$
\EndFor
\State $g_{\textbf{Init}} =$ Hand\_pose\_mapping($g_h$)
\State $\Omega_{\mathcal{O}_i}, \mathcal{O}^{\mathcal{M}_h^k}_i, \mathcal{O}^{\mathcal{A}_h^k}_i$ = $\Phi_{\textbf{DSC}}(\mathcal{O}, \mathcal{O}_i)$
\State $g =$ Grasp\_optimization$(\Omega_{\mathcal{O}_i}, \mathcal{O}^{\mathcal{M}_h^k}_i, \mathcal{O}^{\mathcal{A}_h^k}_i, \mathcal{M}, g_\textbf{Init})$ 
\State $g_{\textbf{final}}=$ Simulator\_refinement$(g)$, add $g_\textbf{final}$ into $\mathcal{G}$
\EndFor

\end{algorithmic}
\end{algorithm}

\subsection{Knuckle-level Hand-Object Contact\label{sec:knuckle_level_contact}}
\begin{figure}
    \centering
    \includegraphics[width=0.95\linewidth]{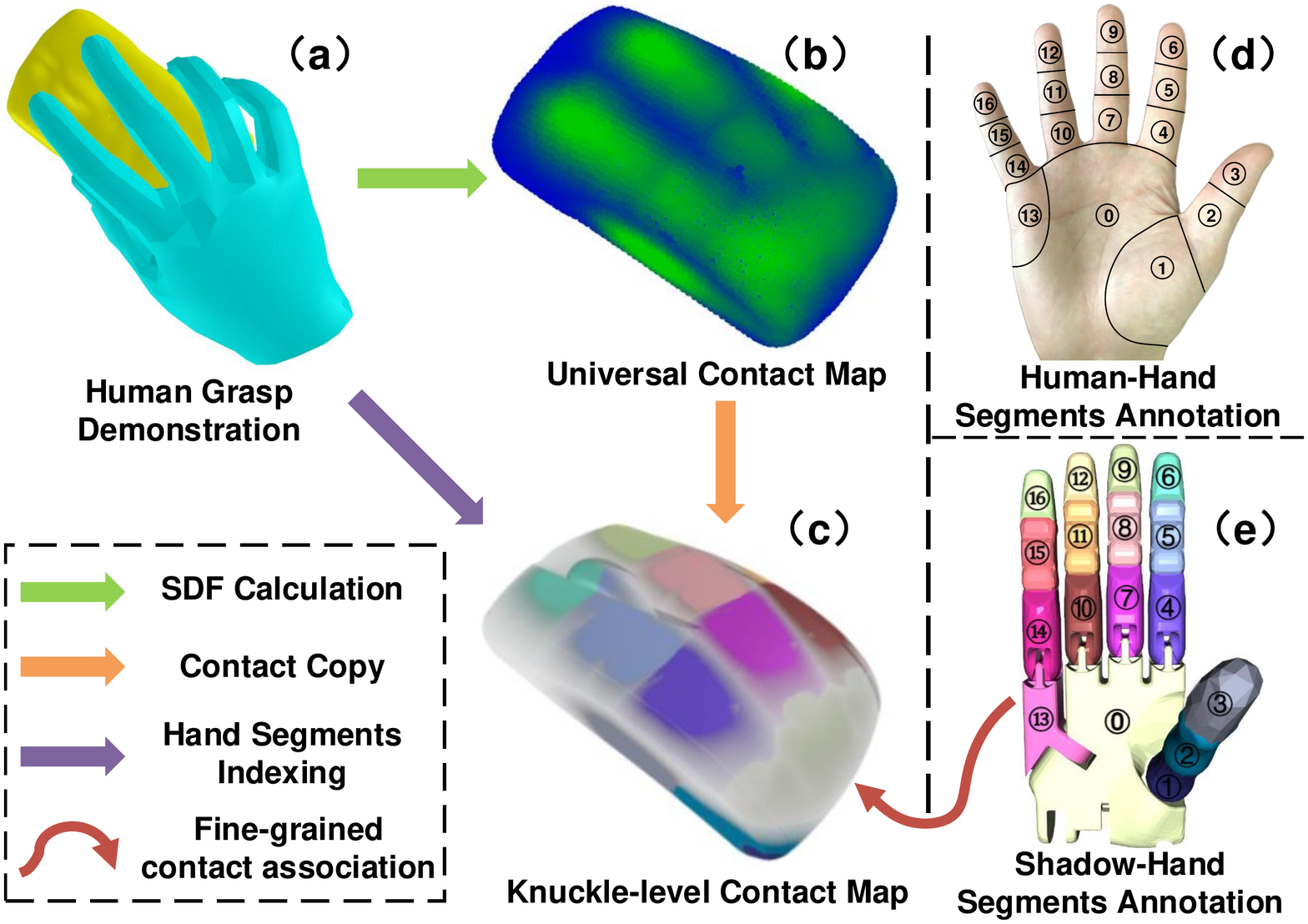}
    \caption{Illustration of  knuckle-level hand-object contact modeling through a human hand grasp demonstration (functional grasp on a mouse). Best view in color and zoom in.}
    \label{fig:knuckle_level_contact}
\end{figure}

Given a human grasp demonstration, shown in Fig. \ref{fig:knuckle_level_contact}(a), the contact map $\Omega_{\mathcal{O}}$ on object surface is digitized as:
\begin{equation}
\begin{aligned}
    d(\mathcal{P}^{\mathcal{O}}, {\mathcal{M}_h}) &= max(0,  SDF(\mathcal{P}^{\mathcal{O}} | \mathcal{M}_h)) \\
    \Omega_\mathcal{O} &= 1 - 2 \cdot(\mathbb{S}(2 \cdot d(\mathcal{P}^{\mathcal{O}}, {\mathcal{M}_h})) - 0.5) 
\end{aligned}
\label{equ:contact_object}
\end{equation}
where $\mathbb{S}(\cdot)$ denotes for Sigmoid activation function. $\mathcal{P}^{\mathcal{O}} \in \mathbb{R}^{N \times 3}$ denotes the uniformly sampled points on the object surface. $SDF(\cdot | \cdot)$ denotes the Signed Distance Field function, the magnitude of the output represents the distance to the given surface model, the sign indicates whether the region is inside (-) or outside (+). Considering that the demonstrated human hand $\mathcal{H}_h$ should not interpenetrate with the object $\mathcal{O}$, a numerical truncation is performed. The resulting contact map $\Omega_\mathcal{O}$ shown in Fig. \ref{fig:knuckle_level_contact}(b) is normalized into range [0, 1]. The points on object surface that are close enough to the hand mesh is denoted as $\mathcal{O}^c$, shown in the green points of Fig. \ref{fig:knuckle_level_contact}(b).

As illustrated in Fig. \ref{fig:knuckle_level_contact}(d), the human hand comprises 17 distinct regions, commonly referred to as hand segments, based on the kinematic chain of the Shadow Hand, as shown in Fig. \ref{fig:knuckle_level_contact}(e). The knuckle-level contact map $\Omega_\mathcal{O}^{KL}$ presented in Fig. \ref{fig:knuckle_level_contact}(c) is obtained by identifying which hand segments is closest to the points in the contact map $\Omega_\mathcal{O}$. Those points in $\mathcal{O}^c$ that are closet to the $i$-th hand segment $\mathcal{M}_h^i$ is denoted as $\mathcal{O}^{\mathcal{M}_h^{i}}$. Corresponding areas of the knuckle-level contact map  $\Omega_\mathcal{O}^{KL}$ are annotated with the same color as the hand segments of the Shadow Hand. 

The knuckle-level contact map provides information on which hand segments should make contact with which areas of the object surface. This information is used in the subsequent steps of the functional grasp synthesis algorithm to optimize the grasp based on fine-grained hand-object contact objectives.

Moreover, contact map on human hand segments $\mathcal{M}_h$ is calculated as following:
\begin{equation}
\begin{aligned}
    d(\mathcal{P}^{\mathcal{M}_h^{i}}, \mathcal{O}) &= max(0,  SDF(\mathcal{P}^{\mathcal{M}_h^{i}} | \mathcal{O})) \\
    \Omega_{\mathcal{M}_h^{i}} &= 1 - 2 \cdot(\mathbb{S}(2 \cdot d(\mathcal{P}^{\mathcal{M}_h^{i}}, {\mathcal{O}})) - 0.5) \\
        \Omega_{\mathcal{M}_h} &= \bigoplus_{i=1}^{N} \Omega_{\mathcal{M}_h^{i}}
\end{aligned}
\end{equation}
where $\mathcal{P}^{\mathcal{M}_h^{i}}$ denotes sampled points  on the $i$-th region of human hand, $\Omega_{\mathcal{M}_h^{i}}$ denotes the normalized contact map of the $i$-th region, $N$ is the number of hand segments. The complete contact map of hand mesh $\Omega_{\mathcal{M}_h}$ is assembled by concatenation operation $\oplus$.  The points on human hand mesh $\mathcal{M}_h$ that is close enough to the object is denoted as $\mathcal{M}_h^{c}$.

\subsection{Auxiliary Anchor Points}
\begin{figure}
    \centering
    \includegraphics[width=0.70\linewidth]{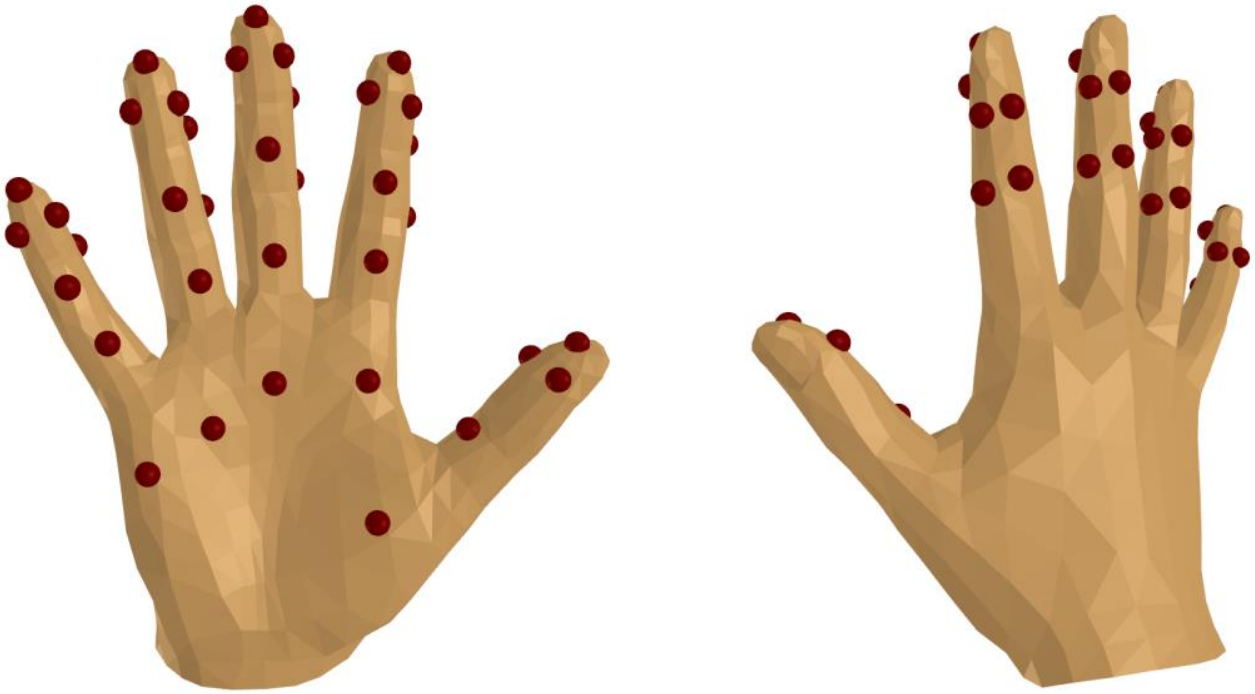}
    \caption{Anchor points distribution of the human hand (in red).}
    \label{fig:mano_anchor_points}
\end{figure}
Although the knuckle-level contact map is useful for identifying fine-grained hand-object contact associations, it is unable to determine the precise location of the contact along the finger segments. To address this limitation, we suggest using auxiliary anchor points to precisely locate hand contacts within a specific finger segments. Specifically, we annotate a total of 41 representative anchor points $\mathcal{A}_{h}=[\mathcal{A}_h^{1}, \mathcal{A}_h^{2}, \ldots, \mathcal{A}_h^{k}], \mathcal{A}_h^{k} \in \mathbb{R}^{3}, k=41$ on the hand mesh $\mathcal{M}_h$. Fig. \ref{fig:mano_anchor_points} demonstrates the distribution of these anchor points $\mathcal{A}_{h}$ on a human hand. Fourteen anchor points are placed on the left and back sides of the distal and middle knuckles, excluding the thumb. These anchor points are selected manually based on the 33  grasp taxonomies introduced in \cite{feix2015grasp}. To determine the contact association between the object contact points $\mathcal{O}^c$ and the anchor points, we use the following formula:
\begin{equation}
\begin{aligned}
        \forall p \in \mathcal{O}^c, \\
        \bm{\pi}_{\mathcal{A}_h}(p) &= \underset{a \in \mathcal{A}_h}{\arg\min}||p - a||  \\
        \Delta_{\mathcal{A}_h}(p) &= ||p -  \bm{\pi}_{\mathcal{A}_h}(p)||_2^{2}
\end{aligned}
\end{equation}
where $\bm{\pi}_{\mathcal{A}_h}(p)$ outputs the nearest anchor point $\mathcal{A}_h^i$ to a given point $p \in \mathcal{O}^c$, $\Delta_{\mathcal{A}_h}(p)$ denotes the $L_2$ projection distance between the point $p$ and its nearest anchor point. The subset of object points that are close enough to anchor point $\mathcal{A}_h^i$ is denoted as $\mathcal{O}^{\mathcal{A}_h^i}$.

Representative anchor points $\mathcal{A}$ are also annotated accordingly on robot hands, as shown in the Appendix.\ref{appendix:anchor_points}.

\subsection{Hand Grasp Mapping\label{sec:hand_grasp_mapping}}
The quality of the initial guess is closely related to the convergence of gradient-based numerical optimization algorithms. For our algorithm, which aims to optimize a human-like robot hand grasp for manipulating a target object, a grasping gesture similar to that of the human hand can serve as a good initialization.

To obtain a human-like grasping gesture for an anthropomorphic hand, it is common to compute hand joint angles using inverse kinematics, given the target cartesian poses of the fingertips. Retargeting the fingertip pose allows for accurate reproduction of point contacts between the robot hand and objects. In addition, direct joint mapping between human and robot hands can help synthesize human-like grasping gestures. To this end, we propose an effective algorithm for mapping human-to-robot hand grasps that balances the above two factors: i) Human-like grasping gesture; ii) Precise fingertip contacts reproduction. The algorithm is formulated as follows:
\begin{equation}
\begin{aligned}
    \mathcal{S}(q_h, q) &= \sum_{i=1}^{I}||\mathbb{F}_i(q_h) - \mathbb{F}_i(q)|| \\
    \mathcal{J}(q_h, q) &= \sum_{j=0}^{J} |q_h^{j} - q^j| \\
    \mathcal{E}(q_h, q) &= \alpha_1 \cdot \mathcal{S}(q_h, q) +  \alpha_2 \cdot \mathcal{J}(q_h, q) \\
\end{aligned}
\end{equation}
where $q_h^j, q^j$ are the $j$-th joint angles of the human hand and the robot hand, respectively, $\mathbb{F}_i(q_h)$ and $\mathbb{F}_i(q) \in \mathbb{R}^{3}$ represent the $i$-th task vector pointing from the fingertip or DIP joint to the origin of the coordinate system. The total cost function $\mathcal{E}$ for grasp pose mapping includes the direct joint $L_1$ mapping error $\mathcal{J}(q_h, q)$ and the fingertips $L_2$ retargeting error $\mathcal{S}(q_h, q)$, where $\alpha_1=1$, $\alpha_2=5$.  The cost function is minimized using the Least-Squares Quadratic Programming algorithm (LSQP) implemented in NLopt Library\cite{nlopt}.

\subsection{Category-level Shape Correspondence}
As we have assumed that objects of the same category share similar geometric structures and functionalities, and that humans tend to grasp them using similar gestures. This leads to hand-object contact consistency when grasping with the intention to use the object. Therefore, if dense shape correspondence is established across category-level objects, it becomes possible to map the hand-object contact between objects of the same category.

To achieve this, we propose establishing dense shape correspondence for category-level objects by assuming that a particular shape can be obtained by deforming a template shape that represents the category. In practice, we propose estimating the deformation field that deforms the shape prior into the shape of of the desired object instance, inspired by\cite{zheng2021deep}. Notably, the neural network is capable of learning a plausible template with accurate correspondences in an unsupervised manner. We trained the neural network from scratch on our self-collected object model dataset. For more details, please refer to the open-source implementation released by \cite{zheng2021deep}.

During network inference phase, each instance shape establishes dense shape correspondence with the template shape, allowing us to map the contact area between a target object and the template object model. We refer to this process as contact diffusion.

 \subsection{Functional Hand Grasp Optimization\label{sec:functional_hand_grasp}}
In order to synthesize grasps for anthropomorphic hands, an intuitive approach is to encourage hand-object contacts to be consistent with human hand grasp demonstration while preserving human-like grasp gesture whenever possible. With this in mind, we propose the following objective functions for functional grasp optimization.
\subsubsection{Contact Consistency}
The contact consistency objective serves two primary purposes: 1) Encouraging contact on both the object surface and the hand mesh to be consistent with the demonstration; and 2) Encouraging pair-wise contact attractions between hand segments and the object while repelling non-pair hand-object contact. It is formulated as follows:
\begin{equation}
\begin{aligned}
\mathcal{L}_{\mathcal{C}} & = |\Omega_\mathcal{O} -\widehat{\Omega_\mathcal{O}}| + |\Omega_\mathcal{M} - \widehat{\Omega_\mathcal{M}}|  \\
& + \lambda_1 \cdot \sum_{i=1}^{N}{dis(\mathcal{M}^{i}, \mathcal{O}^{\mathcal{M}^i})}  \\ 
& - \lambda_2 \cdot \sum_{i=1}^{N}\sum_{j\neq i}^{N}\max(dis(\mathcal{M}^{i}, \mathcal{O}^{\mathcal{M}^j}), d_1)
\end{aligned}
\end{equation}
where $dis(\mathcal{M}^i, \mathcal{O}^{\mathcal{M}^i})$ outputs the Euclidean distance between the $i$-th link of the hand mesh $\mathcal{M}^i$ and its corresponding contact area $\mathcal{O}^{\mathcal{M}^i}$ on the object surface. $d_1=2.5$ cm is used for repelling loss truncation. 

\subsubsection{Anchor Points Alignment} As aforementioned, auxiliary anchor points on the hand are used to locate precise contacts on finger segments. To align precise hand-object contacts with anthropomorphic hands, the loss is formulated as follows:
\begin{equation}
\begin{aligned}
    \mathbb{I}(\mathcal{A}^i, \mathcal{O}^{\mathcal{A}^{i}}) &=
    \begin{cases}
     1 & \text{if} \quad \min(dis(\mathcal{A}^i, \mathcal{O}^{\mathcal{A}^{i}})) > d_2, \\
     0 & \text{otherwise}. 
    \end{cases} \\
    \mathcal{L}_{\mathcal{A}} &= \sum_{i=1}^{41}\mathbb{I}(\mathcal{A}^i, \mathcal{O}^{\mathcal{A}^{i}}) \cdot {dis(\mathcal{A}^i, \mathcal{O}^{\mathcal{A}^i})}
\end{aligned}
\end{equation}
where $dis(\mathcal{A}^i, \mathcal{O}^{\mathcal{A}^i})$ outputs the Euclidean distance between the $i$-th anchor point and its corresponding contact area $\mathcal{O}^{\mathcal{A}^i}$. The  indicator function $\mathbb{I}$ that determines whether an anchor point should be activated based on the comparison between the shortest Euclidean distance and a threshold $d_2=1$ cm.

\subsubsection{Hand Gesture Consistency}
This term regularizes the optimized grasp should be close to the initial grasp, and is formulated as follows:
\begin{equation}
    \mathcal{L}_\mathcal{G} = \lambda_3 \cdot |q - \hat{q}| + \lambda_4 \cdot |H_T - \hat{H_T}| + \lambda_5 \cdot\mathcal{D}(H_R, \hat{H_R})
\end{equation}
where$H_T$ and $H_R$ denote the translation and rotation of the hand, respectively. $\mathcal{D}(H_R, \hat{H_R})$ computes the rotation distance between two quaternion representations.

\subsubsection{Hand-Object Interpenetration}
In order to generate physically plausible hand grasp, interpenetration between the hand and the object is penalized as follows:
\begin{equation}
    \mathcal{L}_{\mathcal{IP}} = \lambda_6 \cdot \sum \min{(-\langle \bm{1}, SDF(\mathcal{P}^{\mathcal{O}}|\mathcal{M})\rangle, 0)}
\end{equation}
where $\bf{1}$ is a 2D one-vector, and $\langle \cdot, \cdot \rangle$ denotes a dot product. The interpenetration loss $\mathcal{L}_{\mathcal{IP}}$ actually penalize the negative sum of signed distances of the object point to the hand mesh.

\subsubsection{Self Penetration} The self-penetration term penalize the self-collision between hand segments, and it is formulated as follows:
\begin{equation}
    \mathcal{L}_\mathcal{SP} = \lambda_7  \cdot \sum_{i=1}^{N}\sum_{j\neq i}^{N} \min{(-\langle \bm{1}, SDF(\mathcal{P}^{\mathcal{M}^{i}}|\mathcal{M}^{j})\rangle, 0)}
\end{equation}

Finally, the overall objective function for functional grasp optimization combines all the aforementioned objectives:
\begin{equation}
    \mathcal{L} = \mathcal{L}_\mathcal{C} + \mathcal{L}_\mathcal{A} + \mathcal{L}_\mathcal{G} + \mathcal{L}_\mathcal{IP} + \mathcal{L}_\mathcal{SP}
\end{equation}

The grasp optimization was implemented in Pytorch using the AdamW\cite{adamW} optimizer. Given the hand grasp pose mapped from the human grasp, the algorithm produced reasonable results for over 95\% of the cases in our self-collected dataset, within 200 steps. The optimization process for each object took around 16 seconds on an RTX-3090 GPU. To ensure the best possible results, we ran the optimization process five times with random seeds and selected the grasp with the lowest loss for each object.

Furthermore, we refined each generated grasp in the the simulator\cite{mujoco} by attempting the grasp. We found this step to be useful in avoiding both intersection between the hand and the object, as well as self-collision between hand fingers. 

Based on the proposed six-step grasp synthesis algorithm, we build a large-scale functional grasping dataset with over 10K grasps on our self-collected object models. Details of our self-collected object is presented in Appendix.\ref{appendix:dataset_colleciton}. 

The hyperparameters for $\lambda_1, \lambda_2, \lambda_3, \lambda_4, \lambda_5, \lambda_6, \lambda_7$ are 5, 2, 5, 5, 2, 1 and 1, respectively.

\section{Dexterous Functional Grasp Generation}
This section presents the Dexterous Functional Grasp Network (DexFG-Net) for functional grasp generation. The overall pipeline is shown in Fig. \ref{fig:framework_overview}, which comprises 3 submodules: Object Reconstruction, Variational Grasp Sampler, and Iterative Grasp Refinement. 

\subsection{Object Reconstruction \label{sec:obj_reconstruction}}
\begin{figure}[t]
    \centering
    \includegraphics[width=1.0\linewidth]{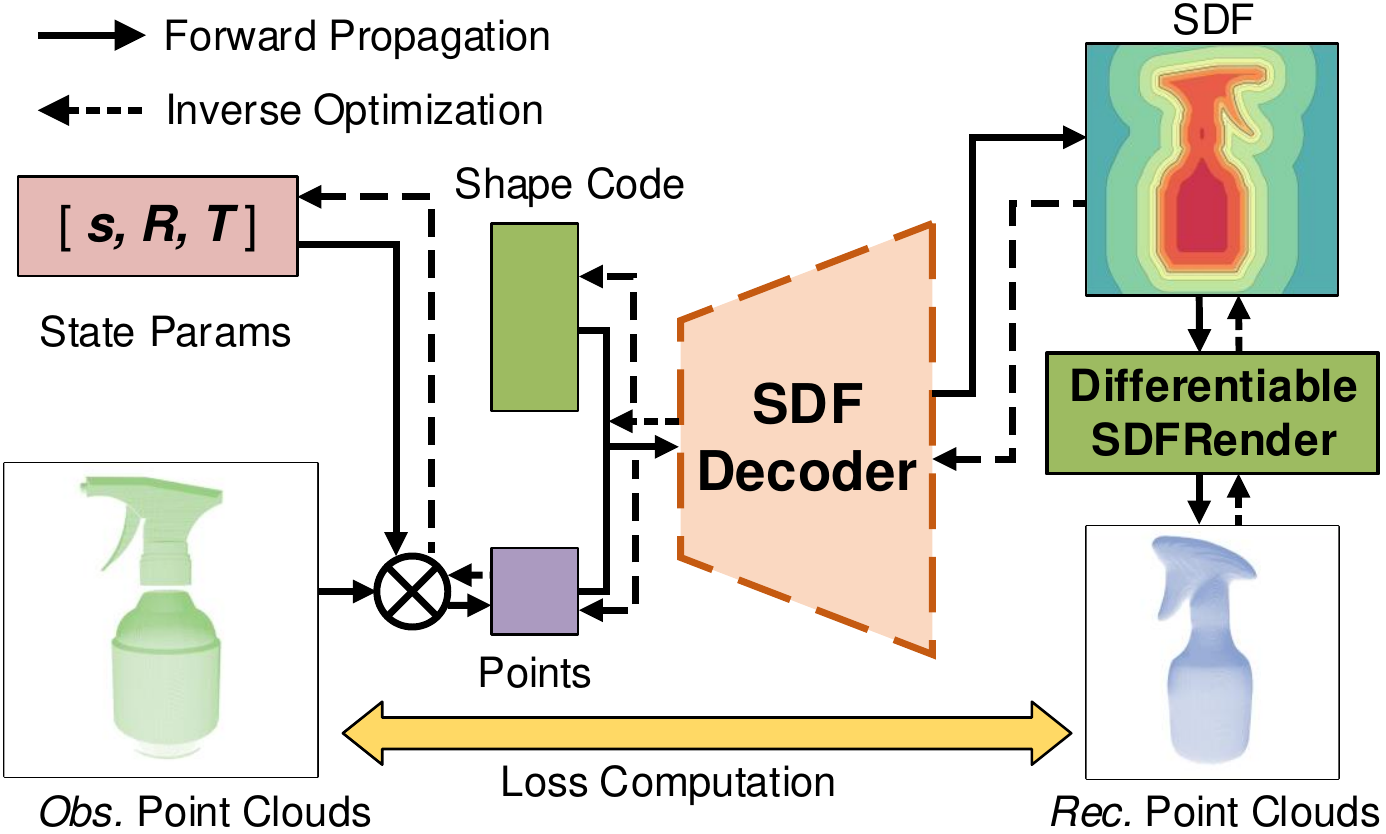}
    \caption{Overall network pipeline for shape completion with partial point cloud.}
    \label{fig:object_reconstruction_pipeline}
\end{figure}

Robotic manipulation with anthropomorphic hands typically involves rich hand-object contacts. Most existing methods assume prior knowledge of the desired object states \cite{fast_planner, contactgrasp}. Some current works propose object reconstruction based on multi-view perception\cite{highdofgrasp,contactpose}. However, using a multi-camera setting imposes certain restrictions for real-world applications.

In this work, we propose to estimate object states from single-view perception. As shown in Fig. \ref{fig:object_reconstruction_pipeline}, our estimation algorithm relies on differentiable rendering and inverse optimization of a shape representation using a deep implicit function. We optimize the shape code and state parameters using gradient descent, given the observed point clouds and shape code of a template object model as input. We use a pre-trained DeepSDF decoder \cite{park2019deepsdf}, with its network parameters kept fixed during object state optimization. The objective function used for optimization is formulated as follows:

1) $\mathcal{L}_{\text{sdf}}$ is the $L_1$ loss that computes the distance between ground-truth SDF value and predicted value. For all observed points on the object surface, the ground-truth SDF values should be zero. Inverse optimization using the DeepSDF decoder requires that the observed points be normalized to the category-level canonical pose and scale for loss computation:
\begin{equation}
    \mathcal{L}_{\text{sdf}} =  \big|\mathrm{SDF}\big({\mathcal{T}_{\mathrm{cls}}(\mathcal{P}_{real}^{\mathcal{O}}, [s, R, T])|\bm{c}}\big)\big|
\end{equation}
where $\mathcal{P}_{real}^{\mathcal{O}}$ denotes the observed points on the object surface, $\mathrm{SDF}(\cdot | \bm{c})$ outputs the predicted signed distance filed of input points, $\bm{c}$ denotes the shape code, $\mathcal{T}_{\mathrm{cls}}(\mathcal{P}_{real}^{\mathcal{O}}, [s, R, T])$ outputs the normalized point cloud based based on the state parameters $[s, R, T]$, $s, R$ and $T$ represents the scale, rotation and translation, respectively. 

2) $\mathcal{L}_{\text{normal}}$ is the similarity loss that computes the cosine similarity between the ground-truth point normals $\bm{n}$ and the estimated normals $\hat{\bm{n}}$:
\begin{equation}
    \mathcal{L}_{\text{normal}} = 1 - \frac{\bm{n} \cdot \hat{\bm{n}}}{max(||\bm{n}|| \cdot ||\hat{\bm{n}}||, \epsilon)}
\end{equation}
where $\epsilon$ is a small value to avoid division by zeros, $\hat{\bm{n}}$ is returned by the Differentiable SDFRender\cite{zakharov2020autolabeling}.

3) $\mathcal{L}_{\text{pc}}$ computes chamfer distance between reconstructed point clouds and observed point clouds of the object surface:
\begin{equation}
\begin{aligned}
    \mathcal{L}_{\text{pc}} &= \frac{1}{|\mathcal{P}_{real}^{\mathcal{O}}|} \sum_{x \in \mathcal{P}_{real}^{\mathcal{O}}} \min _{y \in \mathcal{P}_{rec}^{\mathcal{O}}}\|x-y\|_2^2 \\
    & + \frac{1}{|\mathcal{P}_{rec}^{\mathcal{O}}|} \sum_{y \in \mathcal{P}_{rec}^{\mathcal{O}}} \min _{x \in \mathcal{P}_{real}^{\mathcal{O}}}\|y-x\|_2^2
\end{aligned}
\label{equ:chamfer_distance}
\end{equation}
where $\mathcal{P}_{rec}^{\mathcal{O}}$ is the point clouds reconstructed by the Differentiable SDFRender.

The overall loss for object reconstruction is formulated as follows:
\begin{equation}
    \mathcal{L}_{\textbf{OSE}} = \lambda_\text{sdf} \cdot \mathcal{L}_\text{sdf} + \lambda_\text{normal} \cdot \mathcal{L}_\text{normal} + \lambda_\text{pc} \cdot \mathcal{L}_\text{pc}
\end{equation}
where $\lambda_\text{sdf}, \lambda_\text{normal}$ and $\lambda_\text{pc}$ are 5, 1 and 10, respectively.

The initial state parameters of the object are estimated using the Iterative Closest Point algorithm (ICP). The optimization process was carried out on an RTX-3090 GPU for 300 iterations. The total time required for reconstructing a target object is approximately 15 seconds.

\subsection{Variational Grasp Sampler \label{sec:vgs}}
\begin{figure}
    \centering
    \includegraphics[width=1.0\linewidth]{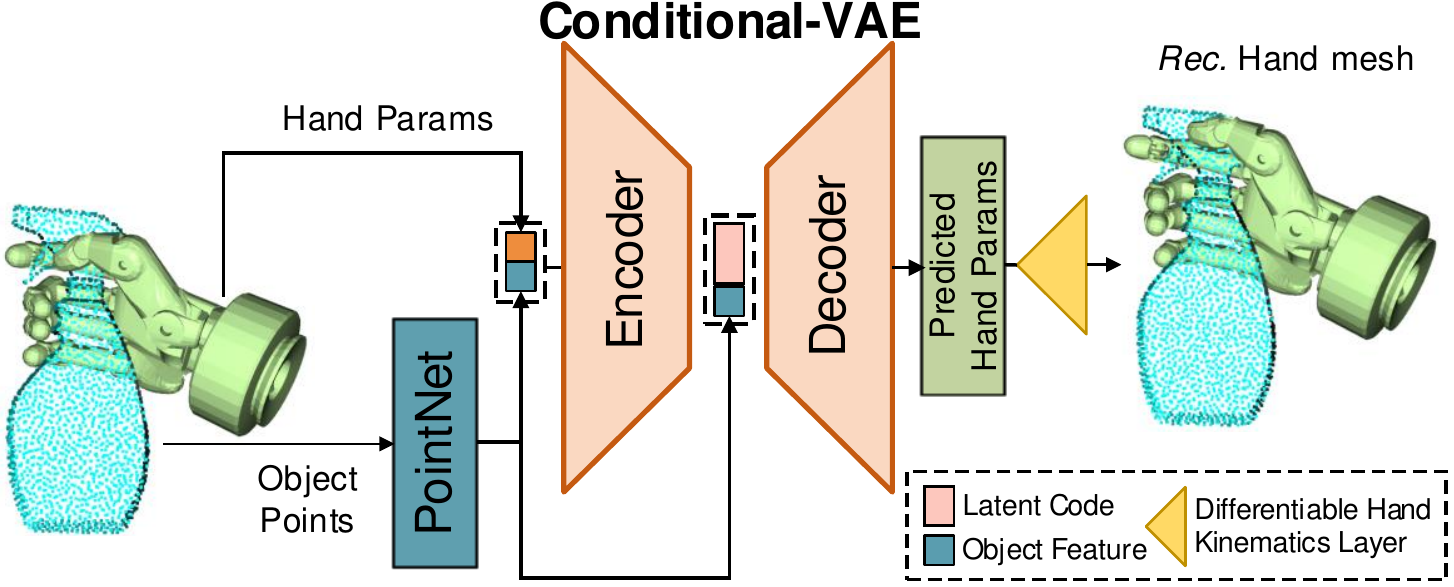}
    \caption{Overall network pipeline for variational grasp sampler based on CVAE.}
    \label{fig:variation_grasp_sampler}
\end{figure}
The Variational Grasp Sampler is based on the Conditional Variational Auto-Encoder (CVAE) and the network pipeline is illustrated in Fig. \ref{fig:variation_grasp_sampler}. Both the encoder and decoder are composed of Multi-Layer Perceptrons (MLP). Our functional grasp dataset was used to train the variational grasp sampler.

During the training state, we use the hand grasp parameters $g$ and object point cloud $\mathcal{P}^{\mathcal{O}}$ as input. PointNet\cite{pointnet} is employed to extract the object feature $\mathcal{F}^{\mathcal{O}}$, while  MLP is used for hand parameters feature extraction $\mathcal{F}^{\mathcal{M}}$. These two features are then concatenated to form the input for the encoder. The encoder learns to map the concated feature to a subspace in the latent space $z$, where $p(z) \sim \mathcal{N}(0, I)$. Given the latent code $z$ and conditional feature $\mathcal{F}^{\mathcal{O}}$, the decoder predicts the hand parameters $\hat{g}$. Given $\hat{g}$ as input, the hand mesh is then reconstructed by a custom differentiable kinematic layer. The objective function for training the network is formulated as follows:

The KL-divergence loss for latent code regularization:
\begin{equation}
     \mathcal{L}_{\mathcal{KL}} = -\lambda_\text{kl} \cdot {\bm{KL}}\left( Q(\bm z|\mathcal{P}^{\mathcal{O}}, g) || \mathcal{N}(0, I)\right)
\label{equation:kl_loss}
\end{equation}

The reconstruction loss for hand model reconstruction:
\begin{equation}
     \mathcal{L}_{\mathcal{R}} = \lambda_\text{v} \cdot \|V - \hat{V}\|^2_2 + \lambda_\text{q} \cdot |q - \hat{q}|
\label{equation:reconstruction_loss}
\end{equation}
where $V$ denotes the hand mesh vertices, $q$ denotes the hand joint angles. The reconstruction loss $\mathcal{L}_{\mathcal{R}}$ consists of the $L_2$ loss for hand vertices reconstruction, $L_1$ loss for joint angles prediction.

The hyperparameters for $\lambda_\text{kl}, \lambda_\text{v}$ and  $\lambda_\text{q}$ are 0.1, 30 and 0.5. Furthermore, the loss functions $\mathcal{L}_{C}$, $\mathcal{L}_{A}$,  $\mathcal{L}_{IP}$ and $\mathcal{L}_{SP}$ proposed in Sec. \ref{sec:functional_hand_grasp} are also used for hand-object interactions reconstruction. 

The overall loss used for variational grasp sampler is stated as follow:
\begin{equation}
    \mathcal{L}_{\textbf{VGS}} = \mathcal{L}_{\mathcal{KL}} + \mathcal{L}_{\mathcal{R}} + \mathcal{L}_\mathcal{C} + \mathcal{L}_\mathcal{A} + \mathcal{L}_\mathcal{IP} + \mathcal{L}_\mathcal{SP}
\end{equation}

During inference state, the encoder is removed, a latent code $z$ is randomly sampled from Normal Gaussian Distribution $\mathcal{N}(0, I)$. Then the latent code $z$ and the conditional object feature $\mathcal{F}^{\mathcal{O}}$ are concatenated to feed into the decoder. The decoder outputs the predicted hand grasp parameters $\hat{g}$.

\begin{figure}
    \centering
    \includegraphics[width=1.0\linewidth]{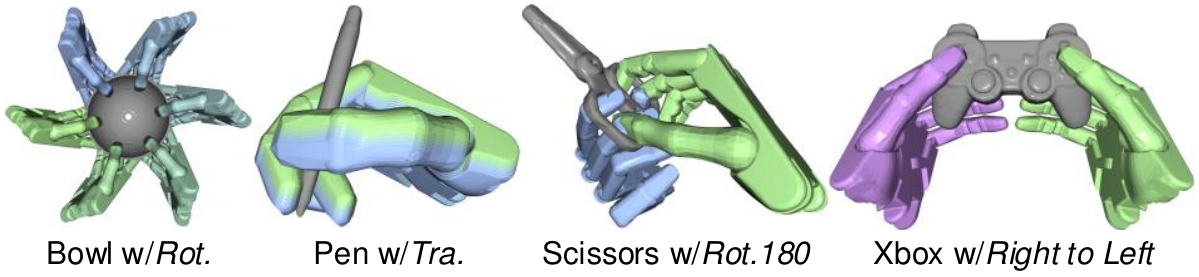}
    \caption{Data augmentation used for training the neural network. Augmentation with rotation and translation are shown in the first three subfigures. The last subfigure shows the grasp mapping of right to left hand, which can be used for grasp generation with left hand.}
    \label{fig:data_augmentation}
\end{figure}

To train the generative-based variational grasp sampler with a diverse set of grasps, we adopt a data augmentation technique. This technique involves rotating and translating the grasp for a target object along the object's symmetric axis while also taking into consideration the functionality constraint. Examples of the data augmentation can be seen in Fig. \ref{fig:data_augmentation}. Additionally, we also introduce limited random noise to the joint angles and hand wrist pose. This helps to introduce some level of disturbance and variability during training, which improves the generalization ability of the network.

To further refine the augmented grasps, they were evaluated in simulation to avoid hand-object inter-penetration and self-penetration between the hand fingers. This helps to ensure that the generated grasps are not only diverse but also feasible and safe for execution on a real robot platform. 





\subsection{Iterative Grasp Refinement \label{sec:igr}}
\begin{figure}
    \centering
    \includegraphics[width=1.0\linewidth]{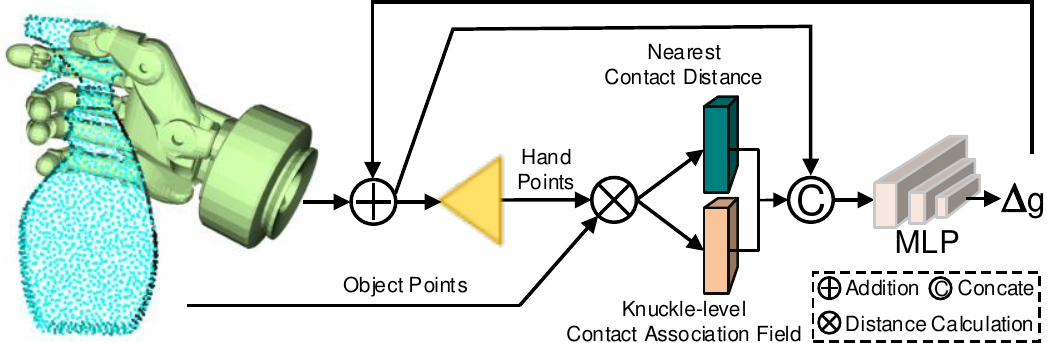}
    \caption{Overall network pipeline for iterative grasp refinement based on contact optimization.}
    \label{fig:refinenet}
\end{figure}

Although the variational grasp sampler can generate human-like grasps, further refinement is necessary to alleviate inter-penetration and self-penetration issues and enhance contact similarity with demonstrated knuckle-level contact. The network architecture is illustrated in Fig. \ref{fig:refinenet}. Specifically, the grasp refinement module takes a sampled grasp $g$ and the sampled points of the target object as input, it calculates both the knuckle-level hand-object contact association filed between the hand segments and the corresponding contact area, as well as the nearest contact distance between the hand and the object. The knuckle-level contact association field and the contact distance are concatenated and fed into the  MLP to predict the residual grasp $\Delta{g}$ transformation.  The loss function is formulated as follows:
\begin{equation}
\begin{aligned}
     \mathcal{L}_{\mathcal{D}} &= \mathcal{D}(g, g^*) \\
     \mathcal{L}_{\textbf{IGR}} &= \mathcal{L}_{\mathcal{D}} + \mathcal{L}_{\mathcal{C}} + \mathcal{L}_{\mathcal{A}}  + \mathcal{L}_{\mathcal{IP}} + \mathcal{L}_{\mathcal{SP}}
\end{aligned}
\label{equation:overall_loss}
\end{equation}
where $\mathcal{L}_{\mathcal{D}}$ regularize the refined grasp $g^*$ should be close to the input grasp $g$.


\section{Experiments}
This section describes extensive experiments carried out to evaluate the performance of the proposed DexFG framework both in simulation and on real robot platform. Following metrics are  used for evaluation.

\textbf{Success Rate} (SR) is a commonly used metric for grasp evaluation. It is important to note that we have employed more stringent criteria for measuring SR in simulation. Specifically, we place objects on the table and reset the generated grasps in the simulator. Then, we increase the flexion of each joint in the hand by an additional +10 degrees towards the palm and lift up the object. Any grasps that are unstable are filtered out by shaking the hand, and only those grasps that consistently hold the object are retained.

\textbf{$\epsilon$-quality} is used to measure the grasp quality, which computes the radius of the largest ball around the origin that fits in the convex hull of the wrench space.

\textbf{Object Displacement} (OD) quantifies the stability of the grasp in simulation\cite{jiang2021hand}. This metric is measured by the average displacement of the object’s center of mass during the grasp simulation, assuming that the hand is fixed and the object is subjected to gravity.

\textbf{Hand Rotation Distance} (HRD) measures rotation distance between the generated Hand wrist pose and ground-truth (GT) wrist pose. For each object category, we annotated the GT wrist pose according to the human hand grasp demonstration. The rotation distance is measured using quaternion as follow:
\begin{equation}
\theta = 2 \arccos (|\langle p, q\rangle|)
\end{equation}
with $\langle p, q\rangle=p_x q_x+p_y q_y+p_z q_z+p_w q_w$.

\textbf{Penetration Depth} and \textbf{Volume} quantify the collision between the hand and the object. To implement these metrics,  we follow the approach used in \cite{jiang2021hand}, where both the hand and objects are voxelized using a grid size of 0.25cm. The penetration depth is measured by the maximum of the distances from hand mesh vertices to the object surface.

\textbf{Functionality Precision} and \textbf{Recall} measure the similarity of contact maps between generated grasps and human grasps. Dense shape correspondence based contact diffusion is used to obtain ground-truth contact maps for category-level objects.

Furthermore, two metrics are used to quantify the performance of shape completion results, the first one is the mean \textbf{Intersection-over-Union} (IoU), which is defined as:
\begin{equation}
    IoU(A, B)=\frac{|A \cap B|}{|A \cup B|}
\end{equation}
where $A$ and $B$ represent the ground-truth and predicted voxel grids for a target object. The ground-truth voxel grid is obtained by meh voxelization. The second metric we employ is the \textbf{Normalized Chamfer Distance} (NCD), which is frequently used in 3-D reconstruction tasks to measure the similarity between two sets of points. To normalize the Chamfer distance, we divide it by the diagonal length of the 3-D object's bounding box.

 \subsection{Experiments for Grasp Synthesis}
This section discusses the experiments carried out to test the functional grasp synthesis algorithm. The goal of these experiments is to showcase the effectiveness and robustness of our proposed six-step grasp synthesis algorithm, which draws inspiration from hand-object contact. To evaluate the grasp performance in simulation, we employed the MuJoCo\cite{mujoco} physics engine.

\subsubsection{Hand Grasp Mapping}
\begin{figure}
    \centering
    \includegraphics[width=1.0\linewidth]{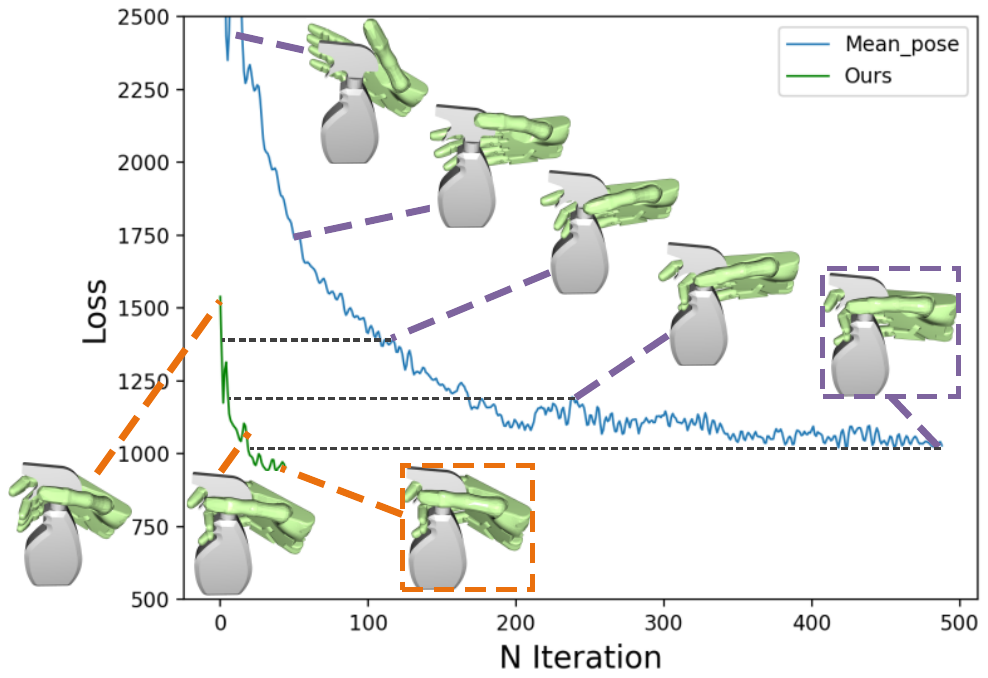}
    \caption{Loss convergence of grasp optimization with mean pose as initialization and our proposed method to obtain initialization. Intermediate cases are also shown for comparison. Best view in color and zoom in.}
    \label{fig:grasp_optimization_process}
\end{figure}
As outlined in Sec.\ref{sec:hand_grasp_mapping}, we propose a hand grasp mapping algorithm that takes into account both fingertip retargeting and direct joint mapping. The resulting hand parameters are then used as the initialization for grasp optimization. Fig. \ref{fig:grasp_optimization_process} presents the convergence of grasp optimization with mean hand pose (denoted in blue) and our proposed method (denoted in green) as the initial grasp parameters. The figure includes several intermediate examples during convergence for each method, with the final convergence results enclosed by a dashed box. Our method provides a superior grasp initialization and converges eight times faster than the mean pose method, as shown by the improved convergence rate in the figure. This highlights the effectiveness of our proposed hand grasp mapping algorithm in generating high-quality grasps in a shorter amount of time.

\begin{figure*}[t]
    \centering
    \includegraphics[width=1.0\linewidth]{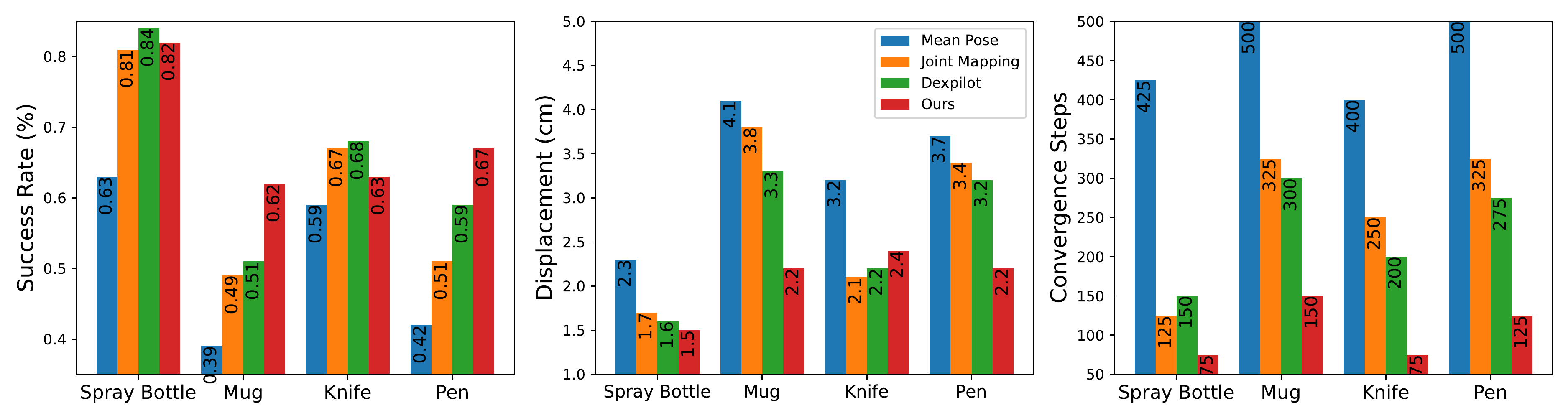}
    \caption{Performance comparison of four different human-to-robot hand grasp mapping methods.}
    \label{fig:hand_grasp_mapping_evaluation}
\end{figure*}

In addition, we conducted a benchmark evaluation of several human-to-robot hand grasp mapping methods on a test dataset consisting of four representative object categories. SR, OD and Convergence Steps (CS) are used as metrics for evaluation. Fig. \ref{fig:hand_grasp_mapping_evaluation} presents the results of the benchmark evaluation, indicating that our proposed method outperforms the other methods in terms of CS, which is on average,  3.3-5.6$\times$ faster than grasp initialization with mean pose, 2.0-2.6$\times$ faster than Dexpilot\cite{handa2020dexpilot}, and 1.7-3.3$\times$ faster than direct joint mapping method.  Furthermore, our method achieves significantly better performance than the other three methods on the mug and pen category, in terms of SR and OD, which is mainly due to our method's ability to synthesize dexterous grasp configurations, such as digging the hand fingers into the mug handles, which undoubtedly make the grasp more stable during the shaking attempts. In the remaining cases, our method achieves slightly better or comparable performance to other baseline methods.

\begin{figure}
    \centering
    \includegraphics[width=1.0\linewidth]{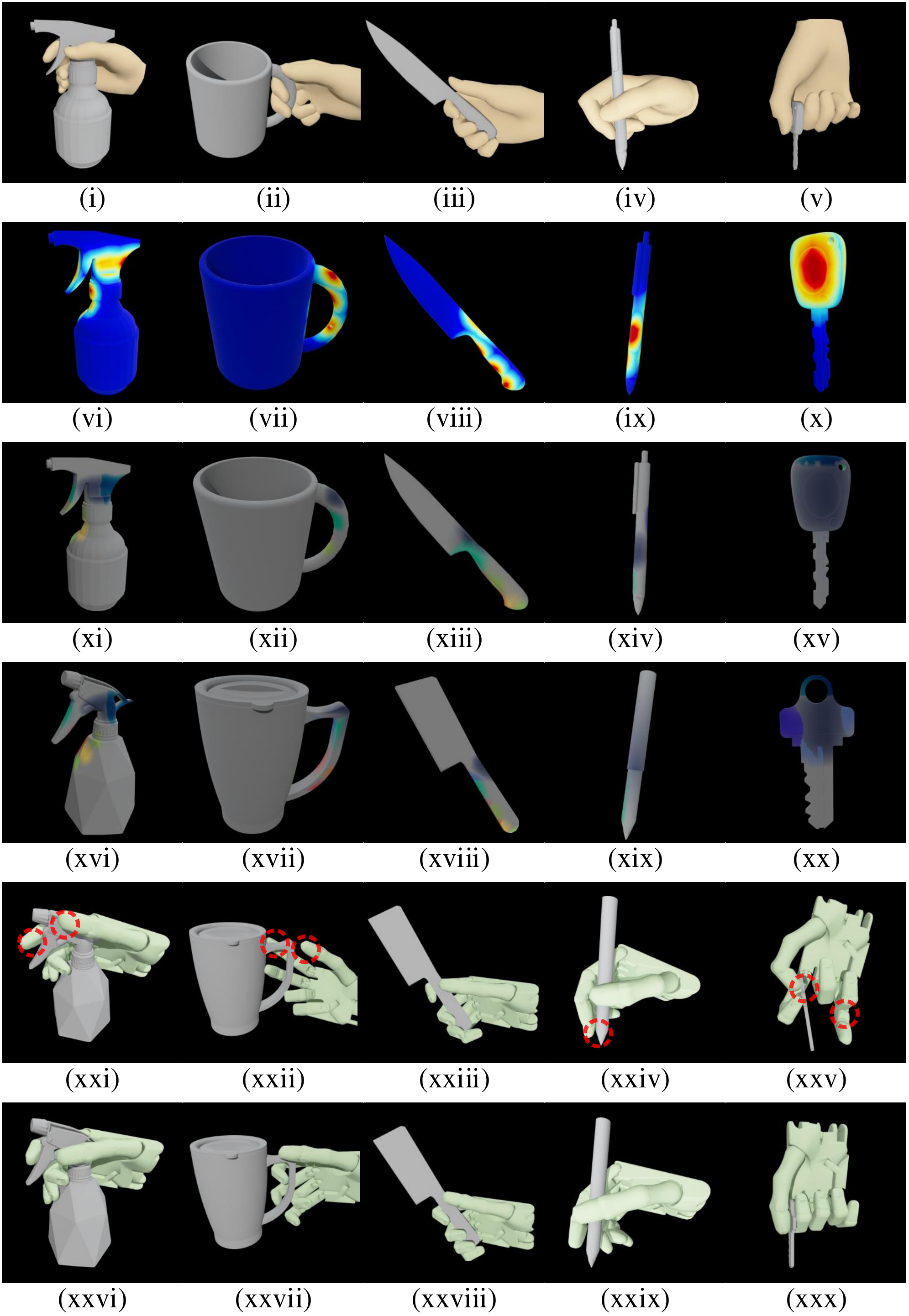}
    \caption{Five cases for grasp synthesis illustration. (i)-(v) Human grasp demonstrations on five objects of different categories. (vi)-(x) Contact map on the objects derived from the corresponding grasp demonstrations. (xi)-(xv) Knuckle-level contact map on the objects derived from the corresponding grasp  demonstrations. (xvi)-(xx) Knuckle-level contact map on other intra-category objects using dense shape correspondence based contact diffusion.  (xxi)-(xxv) Optimized grasps with knuckle-level contact constraint only. (xxvi)-(xxx) Optimized grasps with both knuckle-level contact and anchor points alignment constraints. Unreasonable grasp areas are marked with red dashed circles. Best view in color and zoom in.}
    \label{fig:knuckle_anchor}
\end{figure}
\subsubsection{Fine-grained Contact Modeling} Given a fine-grained contact map for a target object, it is intuitive to optimize the grasp to conform to the contact constraints by ensuring that the hand links touch the associated areas on the object surface. This strategy is effective in generating natural grasp postures for most power grasps that rely on the palm and pad contact\cite{feix2015grasp}. However, it is not robust enough to generate human-like gestures with finger-side contact, such as precision grasps like tripod grasp for holding a pen or opening a bottle cap, or intermediate grasps like lateral grasp for holding a key. 

Fig. \ref{fig:knuckle_anchor} shows a qualitative comparison of synthetic grasps with and without anchor points for five representative object categories, \textit{i.e.} Spray bottle, Mug, Knife, Pen, and Key. Fig. \ref{fig:knuckle_anchor}(i)-(v) shows the human grasp demonstrations, which are manually annotated in GraspIt!. Fig. \ref{fig:knuckle_anchor}(vi)-(x) shows the universal contact maps on the object surface extracted from grasp demonstrations based on signed distance calculation. Fig. \ref{fig:knuckle_anchor}(xi)-(xv) shows the fine-grained knuckle-level contact maps for comparison with the general contact maps, which encode rich contact correspondence between the activated contact area and the hand segments. Fig. \ref{fig:knuckle_anchor}(xxi)-(xxv) shows the diffused knuckle-level contact maps for another intra-category object using dense shape correspondence; Fig. \ref{fig:knuckle_anchor}(xxi)-(xxv) shows the optimized grasp based on knuckle-level contact constraint, indicating that explicitly stating the hand-object contact relationship makes the optimized results look human-like.  Fig. \ref{fig:knuckle_anchor}(xxvi)-(xxx) shows the optimized grasp based on knuckle-level contact constraint and anchor point alignment, revealing that the addition of auxiliary anchor point alignment makes the results more in line with human hand grasp habits.

Furthermore, we conducted quantitative experiments to demonstrate the effectiveness of anchor points alignment. Tab.\ref{tab:anchor_points_alignment} presents a comparison between grasp optimization with and without anchor points. Physical refinement in simulation is not applied for fair comparison. As expected, the optimization results with anchor points alignment achieves significantly higher functionality performance, particularly for the mug, pen, and key object categories, since side contact is required to generate human-like grasp gestures for these categories. Synthetic grasps with additional anchor alignment achieves 21.8\%, 8.5\% and  14.6\% precision improvement for the mug, pen and key categories, and 41.7\%,  18.0\% and  20.6\% recall improvement, respectively. Moreover, it also achieves slightly better performance for spray bottle and knife categories with power grasp taxonomies.
\begin{table}[]
\caption{Functional contact consistency evaluation. }
\begin{tabular}{c|cc|cc}
\hline
\multirow{2}{*}{} & \multicolumn{2}{c|}{KLC}                       & \multicolumn{2}{c}{KLC+APL}       \\ \cline{2-5} 
                  & \multicolumn{1}{c|}{Precision}  & Recall & \multicolumn{1}{c|}{Precision} & Recall\\ \hline
Spray bottle      & \multicolumn{1}{c|}{0.713}      & 0.828  & \multicolumn{1}{c|}{0.752(+$_{0.039}$)}    & 0.853(+$_{0.025}$)\\
Mug               & \multicolumn{1}{c|}{0.471}      & 0.348  & \multicolumn{1}{c|}{0.689(+$_{0.218}$)}    & 0.765(+$_{0.417}$) \\
Knife             & \multicolumn{1}{c|}{0.737}      & 0.768  & \multicolumn{1}{c|}{0.793(+$_{0.056}$)}    & 0.822(+$_{0.054}$) \\ 
Pen               & \multicolumn{1}{c|}{0.679}      & 0.694  & \multicolumn{1}{c|}{0.764(+$_{0.085}$)}    & 0.874(+$_{0.180}$) \\
Key               & \multicolumn{1}{c|}{0.635}      & 0.723  & \multicolumn{1}{c|}{0.781(+$_{0.146}$)}    & 0.929(+$_{0.206}$) \\\hline
\end{tabular}
\footnotesize{Note, ``KLC" and ``APL" are  short for Knuckle-level Contact and Anchor Points Alignment.}
\label{tab:anchor_points_alignment}
\end{table}

\subsubsection{Correspondence Prediction Evaluation}
Dense shape correspondence mapping is a key component for our grasp synthesis algorithm. It is responsible for contact diffusion across category-level objects, which enables fine-grained contact-based grasp optimization. This technique enables the transition from grasping a single object to grasping objects at the category level. By using the dense correspondence predicted by our method, we can perform keypoint detection on object models within the same category via dense shape mapping between reference and test objects. Hence, we propose to evaluate the accuracy of the correspondence based on keypoint detection. We manually annotated several salient keypoints on four categories of our self-collected object models, \textit{i.e.}, Spray bottle, Mug, Knife, Key, and utilize the percentage of correct keypoints (PCK) as the evaluation metric. 

\begin{figure}[t]
    \centering
    \includegraphics[width=0.95\linewidth]{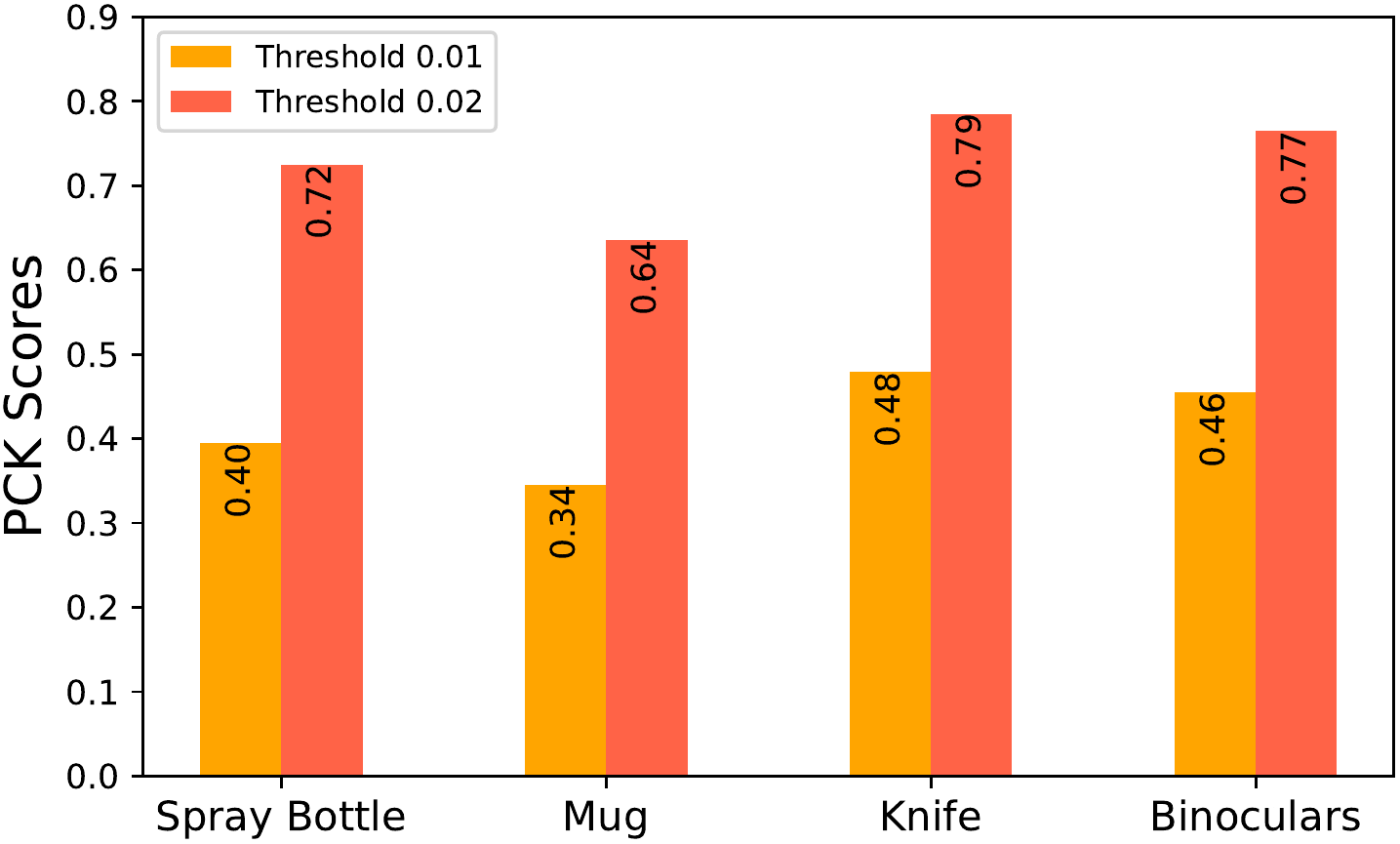}
    \caption{Keypoints detection accuracy using PCK scores with threshold 0.01 and 0.02 on four categories.}
    \label{fig:correspondence_mapping_pck}
\end{figure}

Quantitative results of keypoint detection are shown in Fig. \ref{fig:correspondence_mapping_pck}. In general, our method achieves over 34\% and 64\% PCK accuracy under PCK-0.01 and PCK-0.02 thresholds, respectively, across all four categories. These experimental results demonstrate that our method consistently performs well for all test categories, confirming the effectiveness and robustness of our approach for querying dense shape correspondence through template mapping.

\begin{figure}
    \centering
    \includegraphics[width=1.0\linewidth]{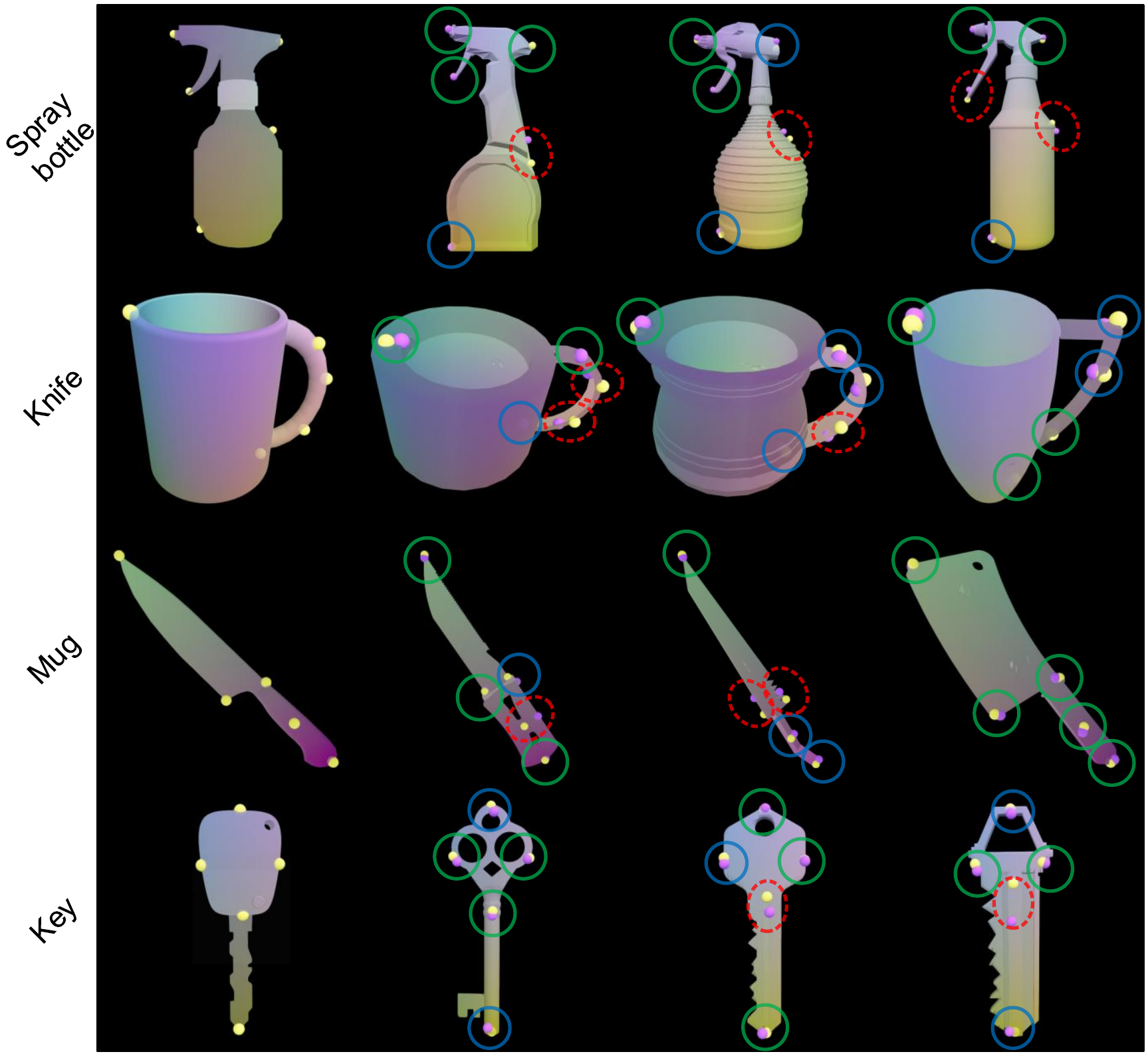}
    \caption{Color-encoded dense shape correspondences and keypoint detection of four representative categories in our self-collected object dataset. Best view in color and zoom in.}
    \label{fig:dense_correspondence}
\end{figure}

Additionally, Fig. \ref{fig:dense_correspondence} presents color-coded demonstrations of dense shape correspondence and keypoint detection of four representative categories. Specifically, for each object category, we manually annotated five keypoints on the object surface. The first column of the figure displays the keypoint locations on the template object model for each category, and the following three columns exhibit the keypoint detection results on three other instances of the same category. Predicted and ground-truth keypoint pairs are enclosed with colorful circles. Keypoint pairs in green circles are prioritized to meet the criteria of the PCK-0.01 metric, and those in blue circles meet the criteria of the PCK-0.02 metric, whereas those in red ellipses do not meet the criteria of the PCK-0.02 metric. These visualization results demonstrate that the method is capable of establishing well aligned dense shape correspondence across between intra-category objects with various shapes. 
 
\subsubsection{Grasp Synthesis with Diverse Robot Hands}
\begin{figure*}[t]
    \centering
    \includegraphics[width=0.95\linewidth]{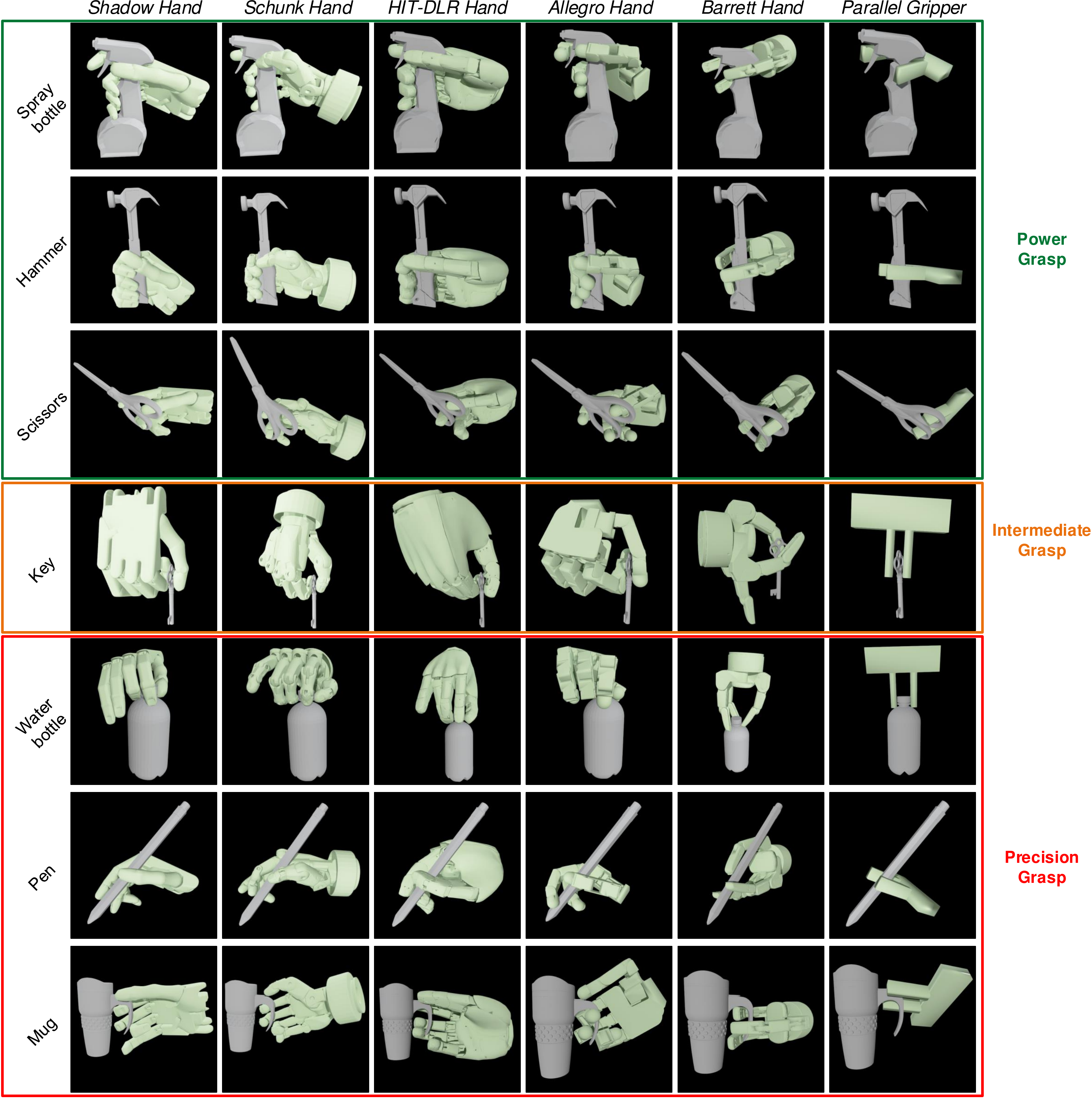}
    \caption{Functional grasp synthesis for tool-use with  six kinematic diverse hand models. Best view in color and zoom in.}
    \label{fig:six_hand_grasp}
\end{figure*}

This subsection presents the generalization ability of our proposed DexFG framework for functional grasp synthesis using a wide range of kinematically diverse hand models. We tested our framework on six different robotic hands: Shadow Hand (22-DoF/22-DoA), HIT-DLR Hand (20-DoF/15-DoA), SchunK Hand (21-DoF/9-DoA), Allegro Hand (16-DoF/16-DoA), Barrett Hand (8-DoF/7-DoA), and Parallel Gripper (2-DoF/1-DoA). DoA stands for Degrees of Actuation, which refers to the number of independent joints that can be controlled. 

We followed the grasp synthesis algorithm described above to synthesize grasps for the six hand models. However, we made some simplifications for robot hands with fewer than five fingers. Specifically, we discarded the little finger for the Allegro Hand and the ring finger and little finger for the Barrett Hand. For the Parallel Gripper, we considered the left finger as the thumb finger, and the right finger could be either the index or middle finger, depending on the specific grasping gesture for the target object. Additionally, we scaled the object mesh by a factor of 1.25 for the Allegro and Barrett hands because their fingers are significantly larger than human hands.

Fig. \ref{fig:six_hand_grasp} depicts the qualitative grasp synthesis results using the six hand models for testing with eight objects from different categories. The proposed functional grasp synthetic pipeline performs consistently well for the majority of the objects and kinematically diverse hand models. Our algorithm can produce dexterous grasp gestures similar to human hands, including power grasps for objects such as spray bottles, hammers, and scissors; intermediate grasps for small cylinders (like cigarettes) and keys; and precision grasps for objects like pens, bottle caps, and mugs. Remarkably, our algorithm can generate dexterous human-like grasps for scissors and mugs that tend to twist the fingers of the hand around the handle, while almost all existing grasp planners have trouble planning these grasp gestures. This is primarily because we employ the articulated hand model to formulate an optimization problem that seeks a hand pose that satisfies the demonstrated hand-object contact constraints. Additionally, the contact-inspired optimization is largely independent of the hand model utilized.

\subsection{Experiments for DexFG-Net In Simulation}

\subsubsection{Shape Completion}
\begin{table}[]
\centering
\caption{Comparison with baseline methods for shape completion on test objects}
\begin{tabular}{c|cc|cc|cc}
\hline
\multirow{2}{*}{Objects} & \multicolumn{2}{c|}{Varley-17} & \multicolumn{2}{c|}{Lundell-19} & \multicolumn{2}{c}{Ours} \\ \cline{2-7} 
                         & IoU           & NCD             & IoU            & NCD             & IoU        & NCD           \\ \hline
Spray Bottle             & 0.502         & 0.050          & 0.528          & 0.043          & \textbf{0.798}      & \textbf{0.028}       \\
Mug                      & 0.523         & 0.047          & 0.548          & 0.039          & \textbf{0.735}      & \textbf{0.027}       \\
Knife                    & 0.513         & 0.037          & 0.544          & 0.033          & \textbf{0.719}      & \textbf{0.013}       \\
Pen                      & 0.498         & 0.044          & 0.514          & 0.041          & \textbf{0.614}      & \textbf{0.026}       \\
Camera                   & 0.559         & 0.056          & 0.598          & 0.048          & \textbf{0.780}      & \textbf{0.037}       \\
Phone                    & 0.573         & 0.046          & 0.602          & 0.035          & \textbf{0.784}      & \textbf{0.026}       \\
Flashlight               & 0.502         & 0.048          & 0.562          & 0.038          & \textbf{0.728}      & \textbf{0.029}       \\
Hammer                   & 0.482         & 0.054          & 0.503          & 0.045          & \textbf{0.587}      & \textbf{0.030}       \\
Scissors                 & 0.449         & 0.049          & 0.480          & 0.040          & \textbf{0.606}      & \textbf{0.027}       \\
Binoculars               & 0.458         & 0.058          & 0.478          & 0.048          & \textbf{0.530}      & \textbf{0.033}       \\ \hline
Average                  & 0.476         & 0.049          & 0.535          & 0.041          & \textbf{0.688}      & \textbf{0.027}       \\ \hline
\end{tabular}
\label{tab:shape_completion}
\end{table}

\begin{figure}
    \centering
    \includegraphics[width=1.0\linewidth]{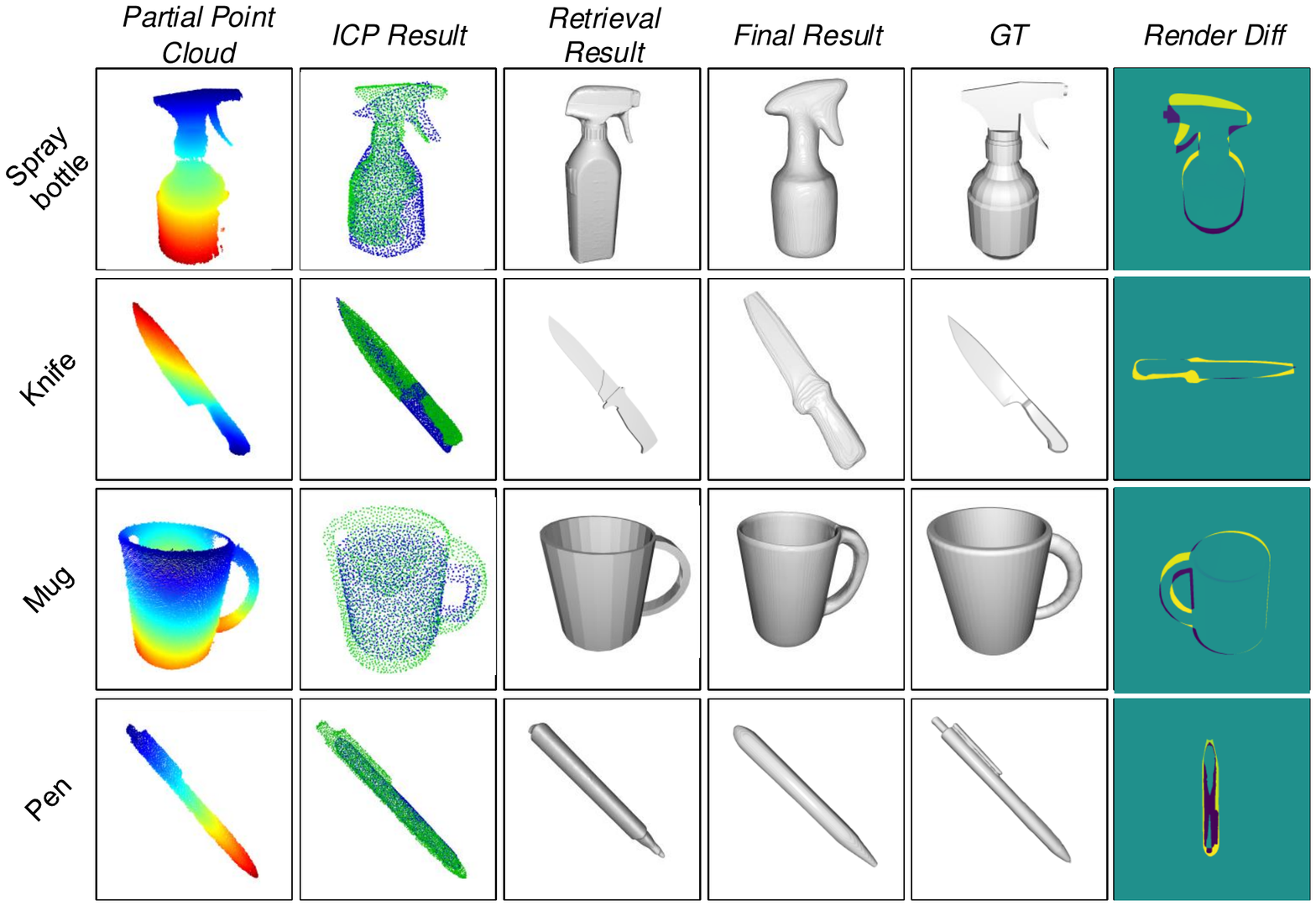}
    \caption{Qualitative results of shape completion for 3-D printed models using partial point cloud captured in real-world scenarios.}
    \label{fig:shape_completion_3d_printed}
\end{figure}

\begin{figure}
    \centering
    \includegraphics[width=1.0\linewidth]{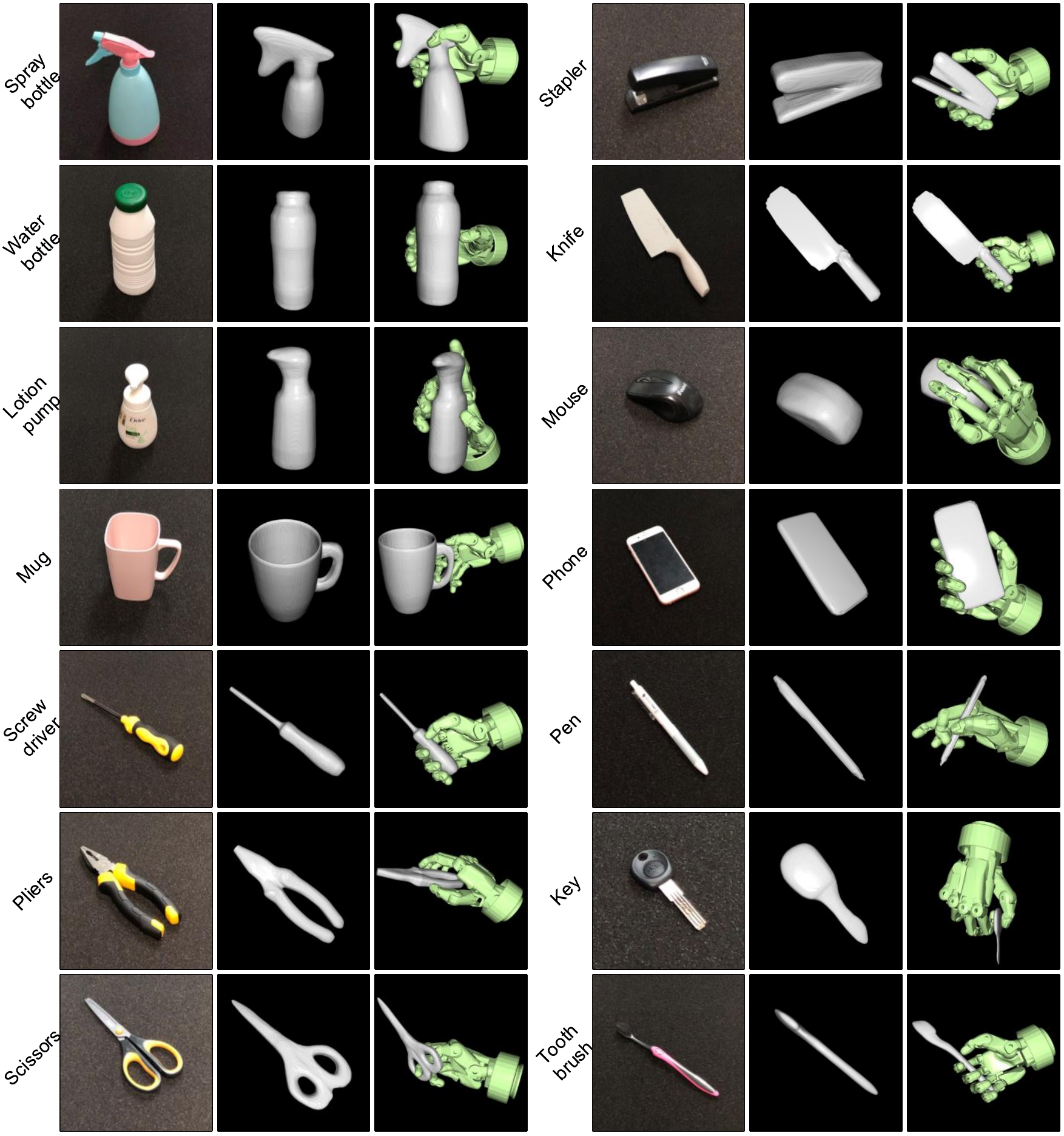}
    \caption{Qualitative results for shape completion and grasp generation for 14 test objects used in real robot experiments. Best view in color and zoom in.}
    \label{fig:real_object_reconstruction_and_grasp}
\end{figure}

To assess the reconstruction performance of our 3-D shape completion module, we compared it with two state-of-the-art shape completion methods for robotic grasping\cite{varley2017shape,lundell2019shape}.  Both of these methods employ a network architecture based on a 3-D Convolutional Neural Network (3-D CNN), which takes in a grid of visible occupied voxel and outputs the shape as a voxel grid. Specifically, the approach proposed by \cite{varley2017shape} predicts voxel grids with a resolution of $32^{3}$, while the one by \cite{lundell2019shape} predicts voxel grids with a resolution of $40^3$. To train these methods, we followed the settings outlined below: 1) For each object in our dataset, we collected 5 partial point clouds captured from different views using Pyrender in simulation; and 2) We adopted the training hyperparameters described in the aforementioned works.


We evaluated all the methods on our object dataset using 10 unseen test objects. Our shape completion algorithm outperformed the baseline methods in terms of higher IoU and lower NCD, as shown in Table \ref{tab:shape_completion}. In contrast to previous works, our method proposes to optimize the object shape based on category-level shape priors using differentiable rendering techniques, while \cite{varley2017shape} and \cite{lundell2019shape} are proposed for universal shape completion without shape priors. We contend that using shape priors can enhance the robustness of shape completion for intra-category objects when only single-view reconstruction is available, particularly in the case of objects with complex shapes such as Spray Bottles, Mugs, and Scissors. However, the methods listed in Tab.\ref{tab:shape_completion}, including our own, do not perform well for certain categories such as scissors, binoculars, and hammer. This could potentially be attributed to the abundance of details presented in these objects. No significant correlation is found between IoU and NCD, as the calculation of NCD involves measuring the distance between points on the surface of two objects, making it challenging to eliminate the influence of shape differences among different object categories. Consequently, NCD is not meaningful for comparing different categories of objects, but it remains a reasonable metric for evaluating reconstruction quality within the same category.

Furthermore, Fig. \ref{fig:shape_completion_3d_printed} presents the qualitative results of our method for shape completion, where the six columns of the figure are arranged in the following order: 1) Partial observed point clouds; 2) Initial results of the ICP algorithm, where a template model was selected for point cloud registration; 3) Retrieval results obtained by computing $L_2$ distance between the intermediate optimized shape code and the shape code database of our object dataset; 4) Optimized final object models using marching cube reconstruction; 5) Ground-truth (GT) object models; and 6) Depth difference maps by subtracting the rendered depth maps of optimized object models from rendered depth maps of the GT object models using Pyrender.


\subsubsection{Functional Grasp Synthesis}
\begin{table*}
\caption{Comparison with Baseline Methods in Terms of Time Cost and Success Rate in Simulation}
\begin{tabular}{l|c|cccccccccc|c}
\hline
Methods      & Time     & \begin{tabular}[c]{@{}c@{}}Spray\\ bottle\end{tabular} & Mug           & Knife         & Pen           & Key           & Camera        & Phone        & Flashlight    & Hammer        & Scissors      & Average        \\ \hline
GPD\cite{ten2017grasp}          & 1.5s     & \textbf{0.90}                                          & \textbf{0.75} & 0.50          & 0.45          & 0.40          & \textbf{0.95} & \textbf{0.9} & 0.85          & 0.80          & 0.45          & 68.5\%$_{(2^{nd})}$          \\
ContactGrasp\cite{contactgrasp} & $>$0.5 hours & 0.85                                               & 0.55          & \textbf{0.70} & 0.40          & 0.30           & 0.65          & 0.75         & 0.60          & 0.65           & 0.40           & 58.5\%$_{(4^{th})}$          \\
Dexpilot\cite{handa2020dexpilot}         & \textbf{0.05s}     & 0.75                                          & 0.50          & 0.40          & 0.25          & 0.45          & 0.80          & 0.85         & \textbf{0.90} & \textbf{0.85} & 0.30          & 60.5\%$_{(3^{rd})}$          \\
Ours         & 0.25s       & 0.80                                                   & 0.60          & 0.60          & \textbf{0.65} & \textbf{0.75} & 0.75          & 0.80         & 0.85           & 0.75          & \textbf{0.55} &\textbf{71.0\%$_{(1^{st})}$} \\ \hline
Ours (Schunk Hand)         & 0.25s       & 0.85                                                   & 0.70          & 0.60          & 0.75 & 0.75 & 0.70          & 0.80         & 0.85           & 0.75          & 0.70 & 75.0\% \\ \hline
\end{tabular}
\footnotesize{Note, the first four rows show grasp performance using the Allegro Hand for fair comparison, the last row shows grasp performance using the Schunk hand for comparison with grasp performance on real robot platform.}
\label{tab:grasp_sr_vs_simulation}
\end{table*}

\begin{table*}[]
\centering
\caption{Comparison with Baseline Methods in Terms of $\epsilon$-quality,  Functionality and Penetration}
\begin{tabular}{lcccccccc}
\hline
\multirow{2}{*}{} & \multirow{2}{*}{$\epsilon$-quality$\uparrow$} & \multirow{2}{*}{\begin{tabular}[c]{@{}c@{}}Hand Rotation\\ Distance (rad)\end{tabular}$\downarrow$} & \multicolumn{2}{c}{Functionality} & \multicolumn{2}{c}{Penetration} & \multicolumn{2}{c}{Self-Penetration} \\ \cline{4-9} 
                  &                                  &                                                                          & Precision$\uparrow$       & Recall$\uparrow$          & Depth (cm)$\downarrow$      & Volume (cm$^3$)$\downarrow$     & Depth (cm)$\downarrow$         & Volume (cm$^3$)$\downarrow$      \\ \hline
GraspIt!\cite{graspit}          & 0.38                             & 1.43                                                                     & 0.310           & 0.349           & 1.92           & 1.79           & \textbf{0.0}      & \textbf{0.0}     \\
ContactGrasp\cite{contactgrasp}      & 0.45                             & 0.87                                                                     & 0.532           & 0.538           & 3.02           & 4.36           & 0.42              & 0.78             \\
DexPilot\cite{handa2020dexpilot}          & \textbf{0.57}                    & \textbf{0.0}                                                            & 0.574           & 0.686           & 3.43           & 6.14           & 0.89              & 1.42             \\
Ours              & 0.42                             & 0.40                                                                     & \textbf{0.702}  & \textbf{0.855}  & \textbf{0.94}  & \textbf{1.68}  & 0.25              & 0.43             \\ \hline
\end{tabular}
\label{tab:functionality_evaluation}
\end{table*}

To demonstrate the effectiveness of our DexFG-Net in generating robust functional grasps, we conducted benchmark grasping trials in simulation. We compared our approach with the following baseline methods: 1) GPD\cite{ten2017grasp} is used to generate a 6-DoF hand wrist pose, assuming a parallel gripper. We adopted the commonly used approach direction sampling method implemented in GraspIt! to sample grasps with a predefined Large Diameter grasp taxonomy used in \cite{feix2015grasp} to approximate parallel-jaw grasping. Grasps that met the frictionless antipodal condition described in \cite{ten2017grasp} are classified as positive grasps, otherwise negative grasps. 2) GraspIt!\cite{graspit} is used to sample 200 grasps in 10 runs based on the built-in simulated annealing planner, and the top 20 grasps with 70k iterations per run were selected based on the Guided Potential Quality Energy cost function. 3) ContactGrasp\cite{contactgrasp} proposed to refine and rank the grasps produced by GraspIt! by optimizing the contact surface to be consistent with a human-demonstrated contact map. 4) Dexpilot\cite{handa2020dexpilot}, which is proposed to map human grasps to robotic hand grasps based on handcrafted rules, we re-implemented the algorithm with the tuned parameters described in \cite{handa2020dexpilot}.

We took the following setup to evaluate the grasp performance in simulation, shown in Tab. \ref{tab:grasp_sr_vs_simulation} and Tab. \ref{tab:functionality_evaluation}: 1) A total of 10 object categories are selected, and 4 objects are sampled in each category for testing; 2) Five grasps are sampled per object for each algorithm; 3) The gripper move freely in simulation without considering motion-planning for robot arm; 4) Complete object mesh models are used for a fair comparison between above methods; 5) To ensure a fair comparison, only the time cost of grasp sampling and refinement is included in the analysis, excluding the time cost of shape reconstruction.

Tab. \ref{tab:grasp_sr_vs_simulation}  presents quantitative comparison between baseline methods and our method, in terms of time cost and grasp success rate. Our method achieves the highest average grasp success performance $71.0\%$,  followed by GPD $68.5\%$, Dexpilot $63.5\%$ and ContactGrasp $55.5\%$, which demonstrates that our DexFG-Net is able to generate stable functional grasps for various object categories. In contrast, although ContactGrasp is also capable of producing functional grasps while it performs the worst mainly due to severe intersection occurs in some of the sampled grasps. It should be noted that our method includes an iterative grasp refinement module for reducing collision\cite{wei2022dvgg}. GPD achieves higher performance for most object categories with cylinder-like or box-like shapes that are suitable for powerful grasp, such as Spray bottle, Mug, Camera and Phone. Our method performs much better for object categories with smaller sizes and requiring side contacts with fingers, such as Pen, Key and Scissors, which demonstrates that the fine-grained contact constraints help to generate stable human-like grasp gestures. Regarding the compute time for each algorithm, Dexpilot achieves the fastest grasp sampling speed, taking about $0.05$ seconds. Our method takes about $0.25$ seconds to sample 5 grasps on a GPU.  GPD takes about $1.5$ seconds to sample a large number of grasps in point cloud with collision filtering and select 5 feasible grasps which meet the force-closure condition. ContactGrasp is much slower than other methods, as it  uses GraspIt! to sample hundreds of grasp candidates that are iteratively refined and reranked such that the contact areas on object surface become closer to the one of the human demonstration.

Tab. \ref{tab:functionality_evaluation} presents quantitative comparison between baseline methods and our method on the test set, in terms of grasp-quality and grasp success rate. As expected, GraspIt! can plan self-penetration free grasps, while it achieves the worst performance in terms of grasp functionality since it is agnostic to contact map. ContactGrasp achieves better functionality performance, since it is designed to refine and rank the grasps sampled by GraspIt! for agreement with human-demonstrated contact map. However, it causes extra inter-penetration and self-penetration. Dexpilot achieves a higher $\epsilon$-quality compared to other methods, while the mapped grasps are actually less realistic due to larger inter-penetration and self-penetration. Dexpilot also shows higher functionality performance compared to ContactGrasp. Our method obtains higher functionality performance and lower inter-penetration compared to other methods, which indicates our method can synthesize physically plausible grasps with higher contact similarity to human-demonstrated grasps. Furthermore, we found that the HRD metric is correlated to the functionality performance, this is mainly due to a similar hand rotation pose is more likely to touch the same area as the demonstrated contact map.

\begin{figure*}
    \centering
    \includegraphics[width=1.0\linewidth]{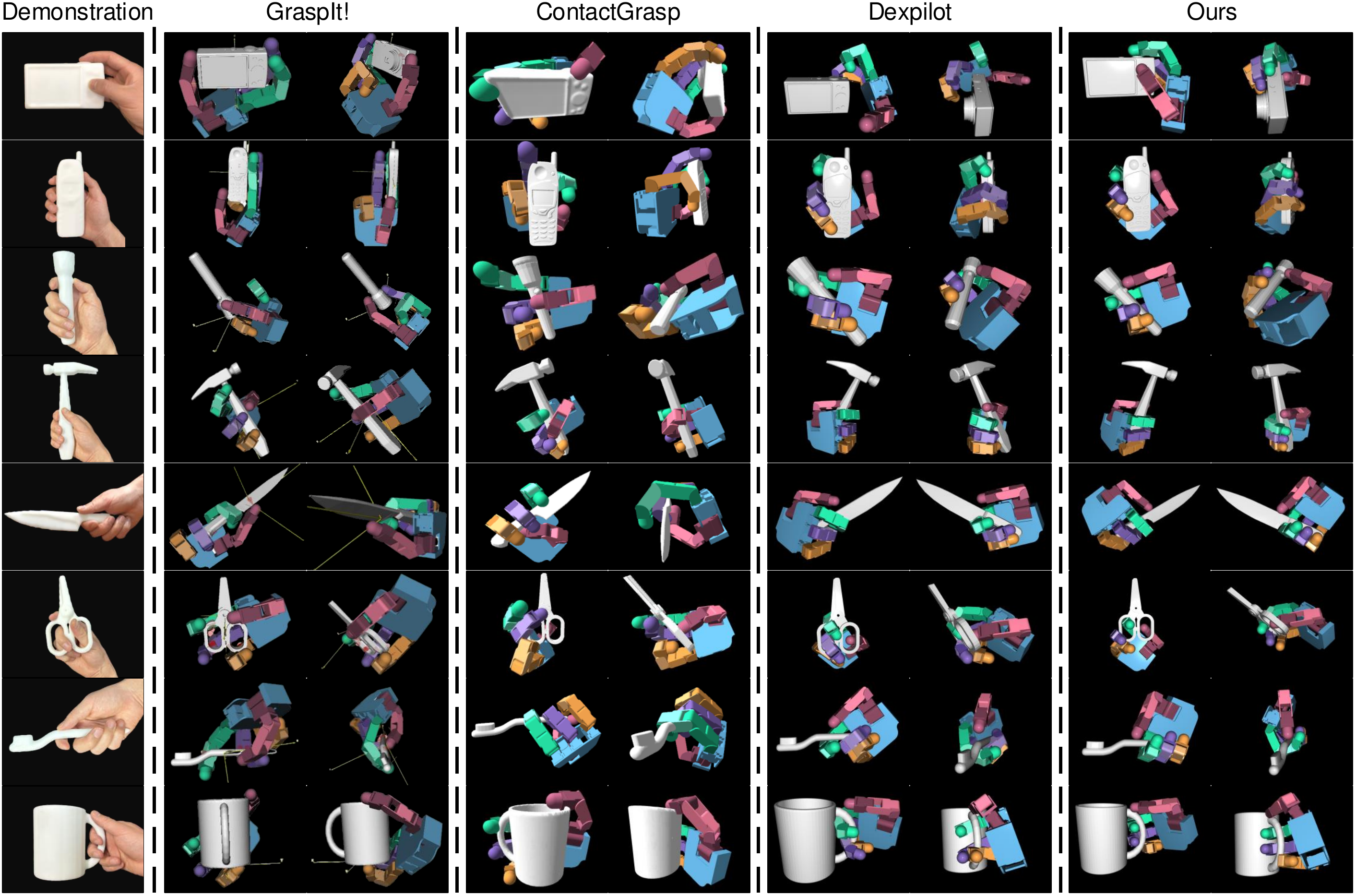}
    \caption{Qualitative comparison of various grasp planner for functional grasp generation given the human hand demonstration using  4-Finger Allegro Hand. Best view in color and zoom in.}
    \label{fig:functional_grasp_planner_vs}
\end{figure*}

Qualitative comparisons with all other three baseline methods are shown in Fig. \ref{fig:functional_grasp_planner_vs}. Our method consistently outperforms the other baseline methods in the vast majority of the presentations shown in both views in the figure. Although Dexpilot can produce human-like grasp gesture, it causes self penetration and interpenetration between the object and gripper for most precision grasp gestures, while our method proposes to avoid collision and maintain human-like grasp gestures. Both statistical and visual comparisons indicate that our generated grasps are significantly closer to human demonstrations than the other baseline methods.

\subsubsection{Ablation Study of Various Loss}
\begin{table}[]
\caption{Ablation Study on Various Loss Functions}
\begin{tabular}{|cl|c|c|c|c|c|}
\hline
\multicolumn{2}{|c|}{Loss Removed}                                     & $\mathcal{L}_\mathcal{C}$   & $\mathcal{L}_\mathcal{A}$   & $\mathcal{L}_\mathcal{SP}$  & $\mathcal{L}_\mathcal{IP}$          & None          \\ \hline
\multicolumn{1}{|c|}{\multirow{2}{*}{Functionality}}    & Precision    & 0.60 & 0.61 & 0.63 & 0.57 & \textbf{0.65}          \\ \cline{2-7} 
\multicolumn{1}{|c|}{}                                  & Recall       & 0.72 & 0.69 & 0.77 & 0.70 & \textbf{0.78}          \\ \hline
\multicolumn{1}{|c|}{\multirow{2}{*}{Penetration}}      & Depth(cm)   & 2.21 & 2.17 & 1.81 & 3.90          & \textbf{1.75} \\ \cline{2-7} 
\multicolumn{1}{|c|}{}                                  & Volume(cm$^3$) & 4.22 & 3.91 & 3.57 & 14.6         & \textbf{3.43} \\ \hline
\multicolumn{1}{|c|}{\multirow{2}{*}{Self-Penetration}} & Depth(cm)   & 0.47 & 0.51 & 1.84 & 0.52          & \textbf{0.44} \\ \cline{2-7} 
\multicolumn{1}{|c|}{}                                  & Volume(cm$^3$) & 0.87 & 0.93 & 4.65 & 1.10          & \textbf{0.79} \\ \hline
\end{tabular}
\label{tab:various_loss}
\end{table}
To study the impact of various loss functions proposed in this article, we conducted an ablation study on these loss terms. As shown in Tab. \ref{tab:various_loss}, we trained the variational grasp sampler with one of these loss terms removed, and no grasp refinement was applied. The model trained with no loss removed is viewed as a baseline. The same experimental setup as in Tab. \ref{tab:grasp_sr_vs_simulation} was used.  As expected, the network trained without inter-penetration loss $\mathcal{L}_{\mathcal{IP}}$ achieves the lowest inter-penetration performance, and the model trained without self-penetration loss $\mathcal{L}_{\mathcal{SP}}$ obtains lowest self-penetration performance. The absence of the above two losses can lead to physically unreasonable grasps. Moreover, the model trained without either knuckle-level hand-object contact loss $\mathcal{L}_{\mathcal{C}}$ or anchor points alignment loss $\mathcal{L}_{\mathcal{A}}$ achieves lower functionality performance compared to the baseline, which indicates stacking these two losses achieve better functionality performance.

\subsubsection{User Study}
\begin{figure*}[t]
    \centering
    \includegraphics[width=.95\linewidth]{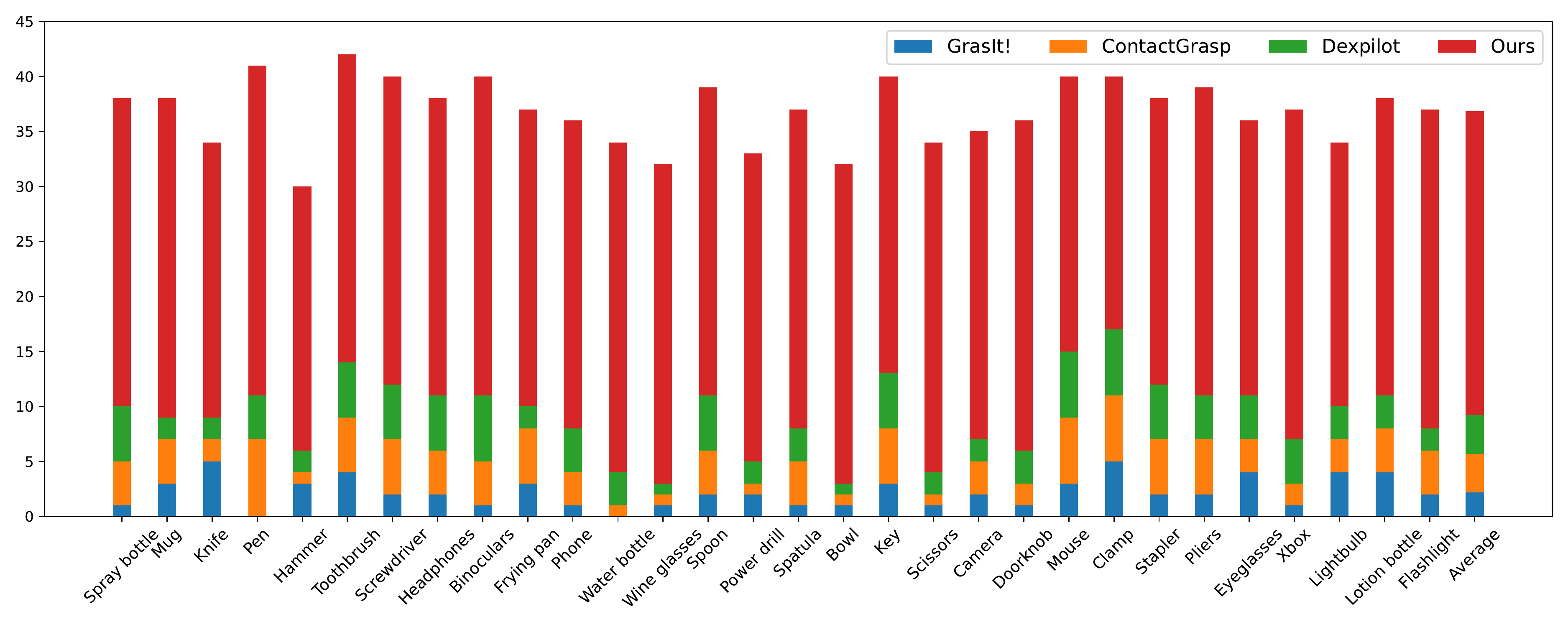}
    \caption{The choice of 30 participants in their preferred grasps for the Allegro Hand, concerning the functionality and shape of the target object. Multiple choices or none can be made, but not encouraged. On average, 74.93\% of the participants considered the grasps produced by our approach closely matched their intuitive choices, compared with 9.59\% of Dexpilot, 9.50\% of ContactGrasp and 5.97\% of GraspIt!.}
    \label{fig:user_study}
\end{figure*}

The goal of this research is to enable robots to perform human-like grasping based on the functionality and shape of the target object. However, the aforementioned metrics do not adequately capture this concept. To quantitatively evaluate each grasp in relation to the functionality and shape of the target object, we introduced a human survey to examine each grasp, as we believe that humans are experts in grasping. A comparative evaluation approach was employed, as quantitatively assessing this similarity is challenging for humans. A representative object was selected from each of the 30 categories in the dataset, and each object was annotated with 1-2 human grasps. For each human grasp, corresponding robotic grasps were generated using three other grasp planners: 1) GraspIt! was used to sample 200 grasps, as mentioned above. However, nearly 95\% of the generated grasps were inconsistent with human grasp habits or unstable. To filter out these grasps, criteria such as approach direction and contact position, as well as the $\epsilon$-metric, were used. The grasp that was most similar to the human demonstration was chosen based on participant votes; 2) The open-source implementation of ContactGrasp\cite{contactgrasp} was utilized to synthesize functional grasps based on a human-demonstrated contact map; 3) Dexpilot \cite{handa2020dexpilot} was employed to transfer human grasp demonstrations to robotic hand grasps. The motion retargeting algorithm was implemented based on the tuned parameters mentioned in their paper. The 4-Finger Allegro Hand was used for fair comparisons with these methods.

Two evaluation criteria were told to human participants: 1) Whether the robots perform human-like grasps; 2) Whether the sampled grasps are in relation to the functionality and shape of objects. Thirty participants took part in the study and were asked to choose their preferred grasps based on these criteria. 

Fig. \ref{fig:user_study} reports the grasp choices by 30 participants in the comparison between baseline methods and our method. On average, $74.93\%$ of the participants selected grasps produced by our method, compared to $5.97\%$ for GraspIt!,  $9.50\%$ for ContactGrasp and  $9.59\%$ for Dexpilot.

\subsection{Experiments for Functional Grasp on Real Robot}
\begin{figure}[t]
    \centering
    \includegraphics[width=0.80\linewidth]{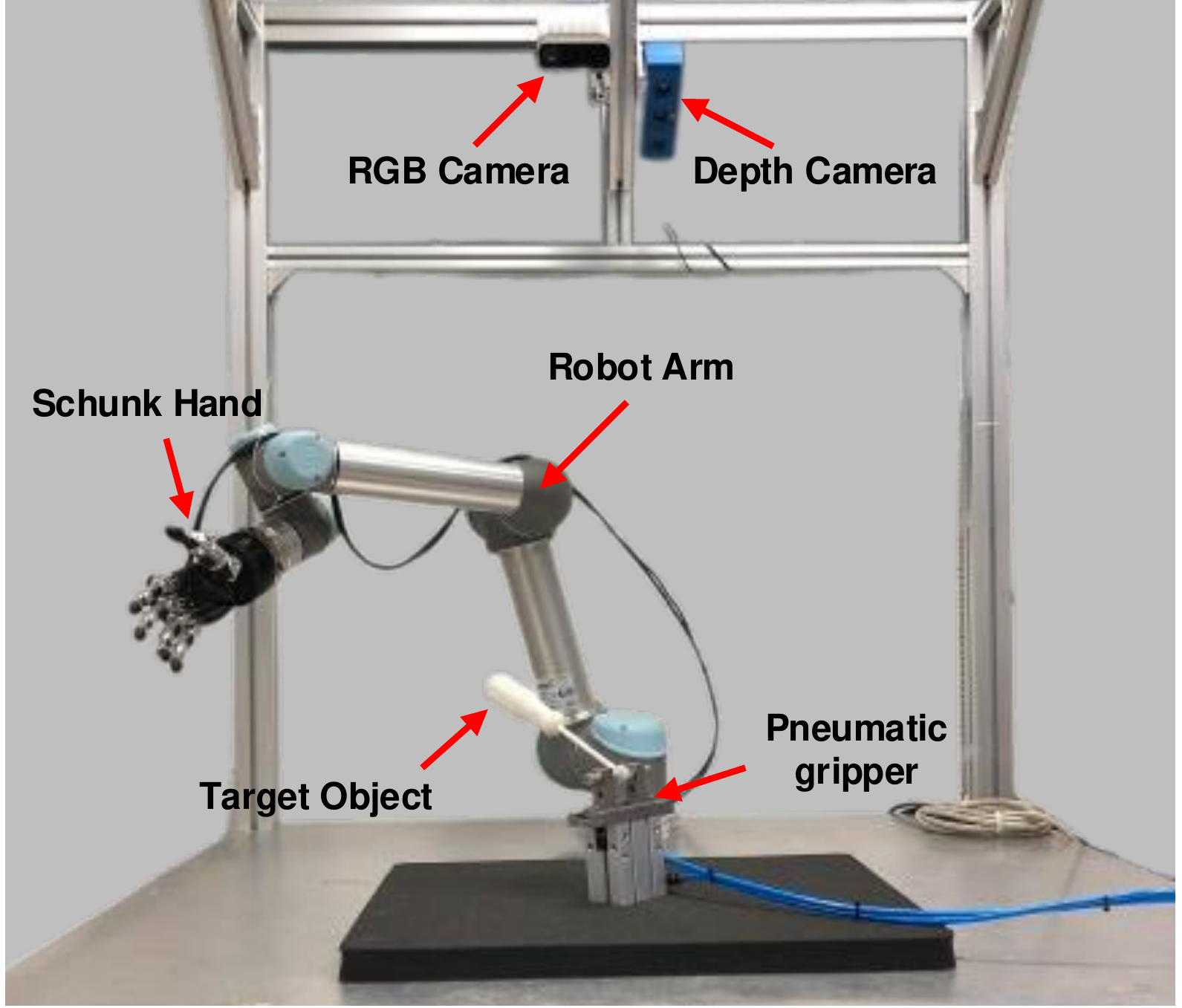}
    \caption{Robotic setup for real robot experiments.}
    \label{fig:real_robot_platform}
\end{figure}

\begin{figure}
    \centering
    \includegraphics[width=1.0\linewidth]{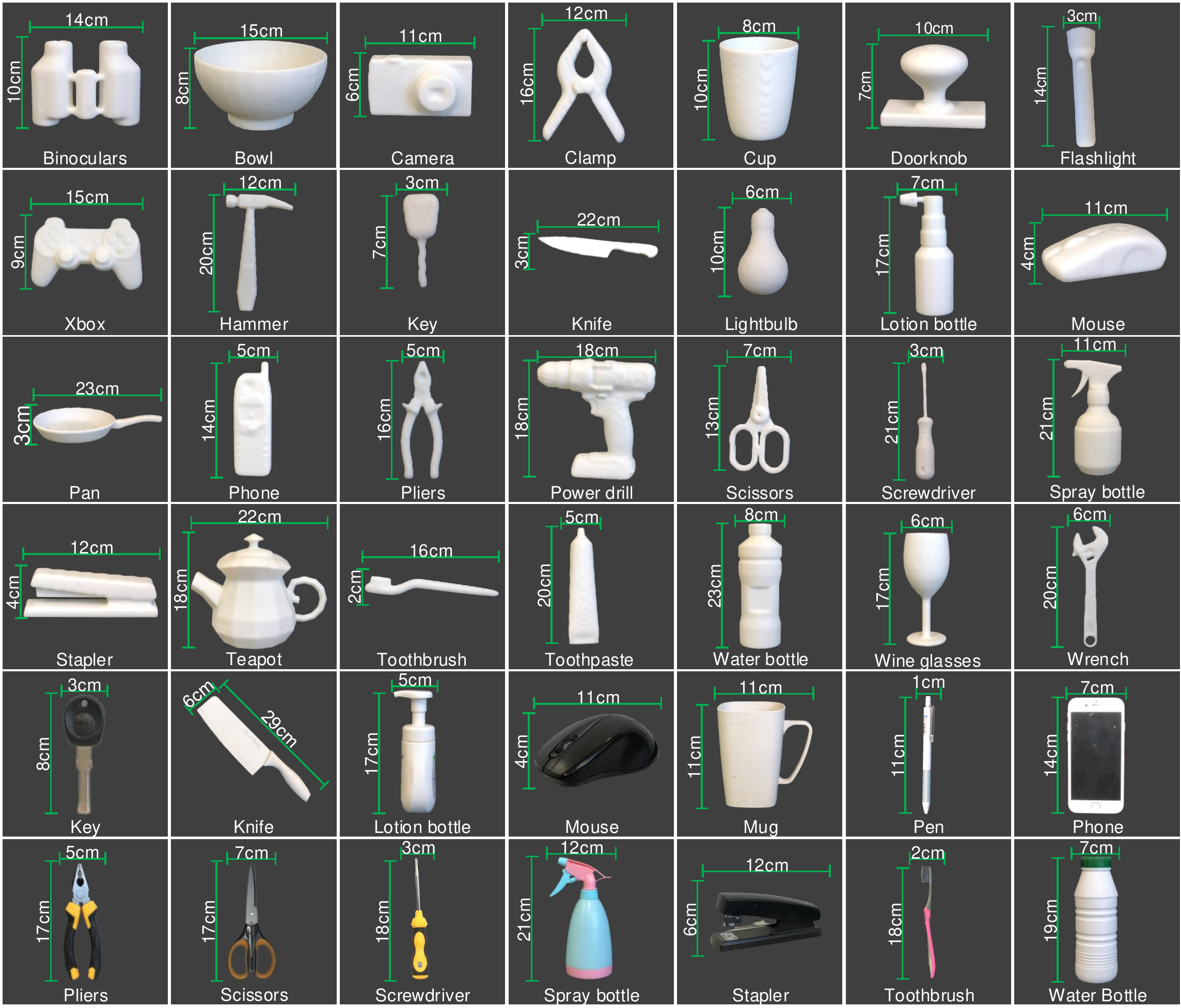}
    \caption{Test objects used in real robot experiments, where 28 3-D printed object (shown in the first 4 rows) collected from YCB dataset and ContactDB dataset and 14 real objects (shown in the last two rows) collected in our lab.}
    \label{fig:real_test_object}
\end{figure}

This subsection presents the performance of the proposed DexFG-Net in real-world scenarios using the robot platform depicted in Fig. \ref{fig:real_robot_platform}. The platform comprises a 6-DoF UR-5 robotic arm and a 5-Finger Schunk Hand. The Kinect Azure and Ensenso N35 camera is statically mounted on top of the platform to capture RGB and Depth images, respectively. The test object set consists of 28 3-D printed objects and 14 real objects collected in our lab as shown in Fig. \ref{fig:real_test_object}. Objects are presented to the robot individually on the tabletop and clamped by a pneumatic gripper if necessary. Foreground object point clouds are segmented by subtracting the background. 

To execute grasps on a real robot, \textit{MoveIt!} integrated in ROS was utilized for motion planning. The obstacle and joint limit constraints are taken as input by \textit{MoveIt!} to plan a collision-free motion path for approaching the target grasp pose. Sometimes, the target grasps can not be executed due to collisions with obstacle or physically unreachable. For a successfully planned grasp, the hand joints move to their desired configuration from an initial configuration. Subsequently, constant torques are applied to the finger joints for 3 seconds. Finally, the robot hand moves upward by 10 cm. A grasp is classified as successful if the robot hand can grasp the target object and lift it to the predefined height without dropping it during translation.

To evaluate grasping performance on a real robot platform, each test object was placed on the table in five different orientations around the $z$-axis, as shown in Fig. \ref{fig:real_robot_platform}. For each scene, the DexFG-Net took the partial point cloud of the scene as input and produces a complete object mesh with ten sampled functional grasps.  All sampled grasps were sorted by descending order with respect to their functionality scores. The first physically reachable grasp was executed to evaluate grasp success rate. The predicted complete object mesh of the 3-D printed object was used to evaluate the shape completion performance.

As shown in Tab. \ref{tab:real_experiments_3d_printed}, 28 3-D printed objects from YCB dataset and ContactDB dataset are used to evaluate both the grasp success rate and shape completion performance. The ground-truth object poses were estimated using ICP algorithm and examined manually. Our method, on average, achieves 79.29\% success rate for 3-D printed objects. In general, Tab. \ref{tab:real_experiments_3d_printed} and Tab.  \ref{tab:grasp_sr_vs_simulation} show a similar trend for those difficult objects, such as the scissors, teapot and toothbrush.  Notably, the average grasp success rate is higher than the result in Tab. \ref{tab:grasp_sr_vs_simulation}, this is mainly due to: 1) The percentage of difficult samples is lower; 2) Simulation experiments take stricter criteria, such as hand shaking after grasp execution; 3) Greater surface friction coefficient of the 3-D printed object. Furthermore, the shape completion module works consistently well on real world scenarios, which demonstrates the robustness of the shape algorithm to sensor noise. 

Tab. \ref{tab:real_lab_objects} reports the grasp performance of 14 real objects. On average, our method achieves a success rate of  68.57\%, which is much lower than the success rate of 3-D printed objects, this is mainly due to: 1) The surface of the real objects are generally smoother, resulting in lower friction coefficients; 2) Sensor noise for objects made of metal and plastic materials, which leads to undesirable shape completion results.




\begin{table}[]
\centering
\caption{Physical functional grasp experiments for 3-D printed models}
\begin{tabular}{c|cc}
\hline
\multirow{2}{*}{Objects} & Grasp   & Shape       \\
                         & Success & Completion (IoU)                                 \\ \hline
Knife                    & 3 / 5   & 0.708                                           \\
Phone                    & 4 / 5   & 0.762                                           \\
Water bottle             & 5 / 5   & 0.756                                           \\
Xbox                     & 4 / 5   & 0.643                                           \\
Doorknob                 & 5 / 5   & 0.766                                           \\
Pan                      & 3 / 5   & 0.684                                           \\
Toothbrush               & 2 / 5   & 0.593                                           \\
Lightbulb                & 5 / 5   & 0.672                                           \\
Wine glasses             & 4 / 5   & 0.624                                           \\
Pliers                   & 4 / 5   & 0.603                                           \\
Camera                   & 4 / 5   & 0.792                                           \\
Toothpaste               & 5 / 5   & 0.735                                           \\
Mouse                    & 4 / 5   & 0.773                                           \\
Teapot                  & 2 / 5   & 0.705                                           \\
Spray bottle             & 4 / 5   & 0.802                                           \\
Key                      & 3 / 5   & 0.712                                           \\
Squeeze bottle           & 5 / 5   & 0.632                                           \\
Power drill                    & 4 / 5   & 0.648                                           \\
Clamp                    & 3 / 5   & 0.652                                           \\
Binoculars               & 5 / 5   & 0.513                                           \\
Bowl                     & 4 / 5   & 0.633                                           \\
Stapler                  & 5 / 5   & 0.689                                           \\
Screwdriver              & 5 / 5   & 0.692                                           \\
Scissors                 & 2 / 5   & 0.621                                           \\
Hammer                   & 5 / 5   & 0.573                                           \\
Flashlight               & 4 / 5   & 0.732                                           \\
Cup                      & 4 / 5   & 0.756                                           \\
Lotion bottle              & 4 / 5   & 0.587                                           \\ \hline
Average                  & 79.29\% & 0.681                                           \\ \hline
\end{tabular}
\label{tab:real_experiments_3d_printed}
\end{table}

\begin{table}[]
\centering
\caption{Physical functional grasp  experiments for real objects}
\begin{tabular}{c|ccc}
\hline
\multirow{2}{*}{Objects} & \multicolumn{1}{c|}{Grasp}    & \multicolumn{1}{c|}{\multirow{2}{*}{Objects}} & Grasp   \\
                         & \multicolumn{1}{c|}{Success}  & \multicolumn{1}{c|}{}                        & Success \\ \hline
Spray bottle             & \multicolumn{1}{c|}{4 / 5}         & \multicolumn{1}{c|}{Stapler}                 &  4 / 5       \\
Water bottle             & \multicolumn{1}{c|}{5 / 5}         & \multicolumn{1}{c|}{Knife}                   &  3 / 5       \\
Lotion bottle              & \multicolumn{1}{c|}{4 / 5}         & \multicolumn{1}{c|}{Mouse}                   &  4 / 5       \\
Mug                      & \multicolumn{1}{c|}{3 / 5}         & \multicolumn{1}{c|}{Phone}                   &  4 / 5       \\
Screwdriver              & \multicolumn{1}{c|}{4 / 5}         & \multicolumn{1}{c|}{Pen}                     &  2 / 5       \\
Pliers                   & \multicolumn{1}{c|}{4 / 5}         & \multicolumn{1}{c|}{Key}                     &  3 / 5       \\
Scissors                 & \multicolumn{1}{c|}{2 / 5}         & \multicolumn{1}{c|}{Toothbrush}              &  2 / 5       \\ \hline
Average                  & \multicolumn{3}{c}{68.57\%}                                   \\ \hline
\end{tabular}
\label{tab:real_lab_objects}
\end{table}

\begin{figure*}[t]
    \centering
    \includegraphics[width=0.90\linewidth]{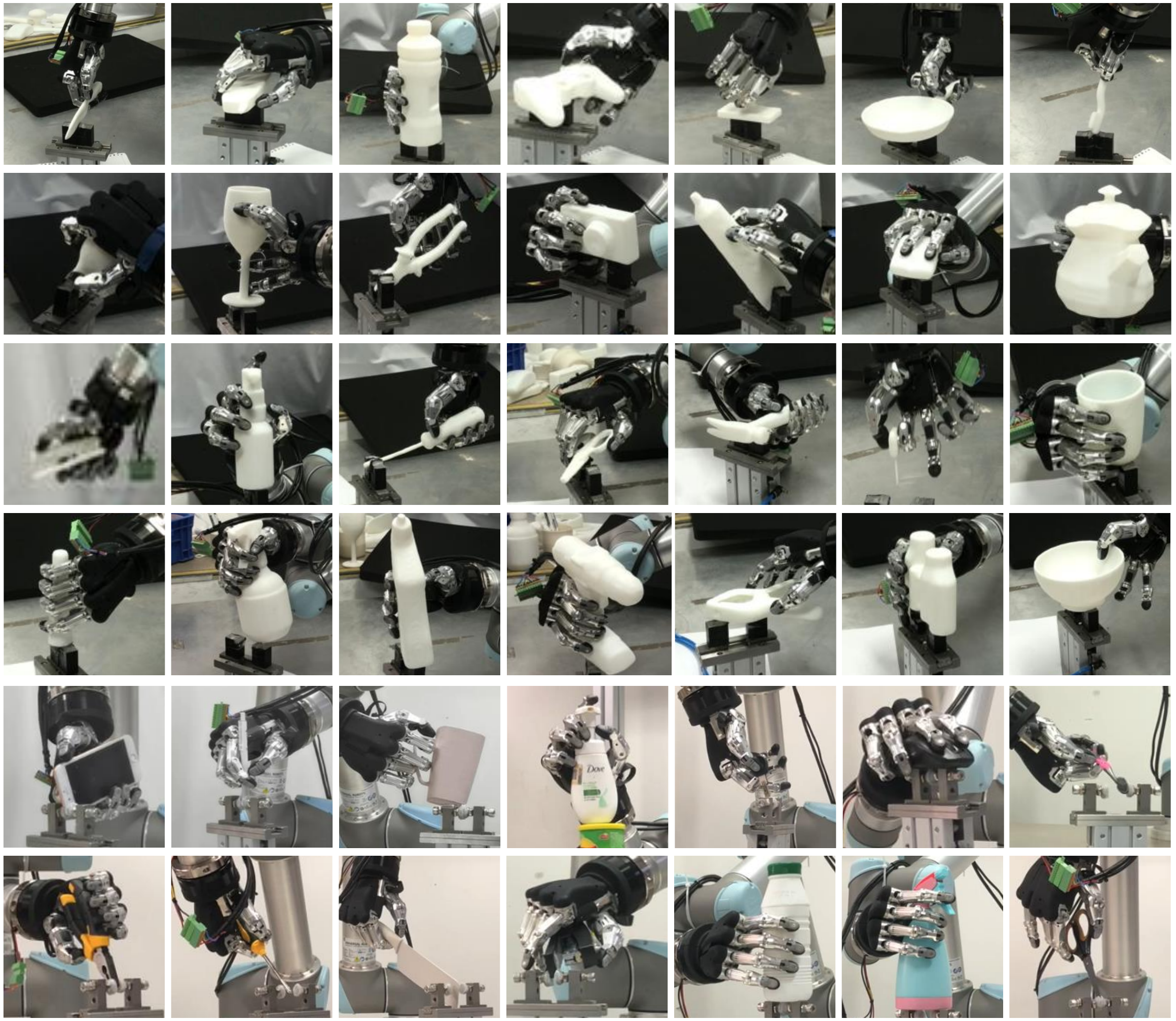}
    \caption{Functional grasps with 5-Finger Schunk Hand on real robot platform. The first four columns show grasps on 28 3-D printed objects, and the last two columns show grasps on 14 real objects collected from our lab. Best view in color and zoom in.}

    
    \label{fig:real_grasps}
\end{figure*}

\subsection{Failed Trials and Limitations}
\begin{figure}[h]
    \centering
    \includegraphics[width=0.85\linewidth]{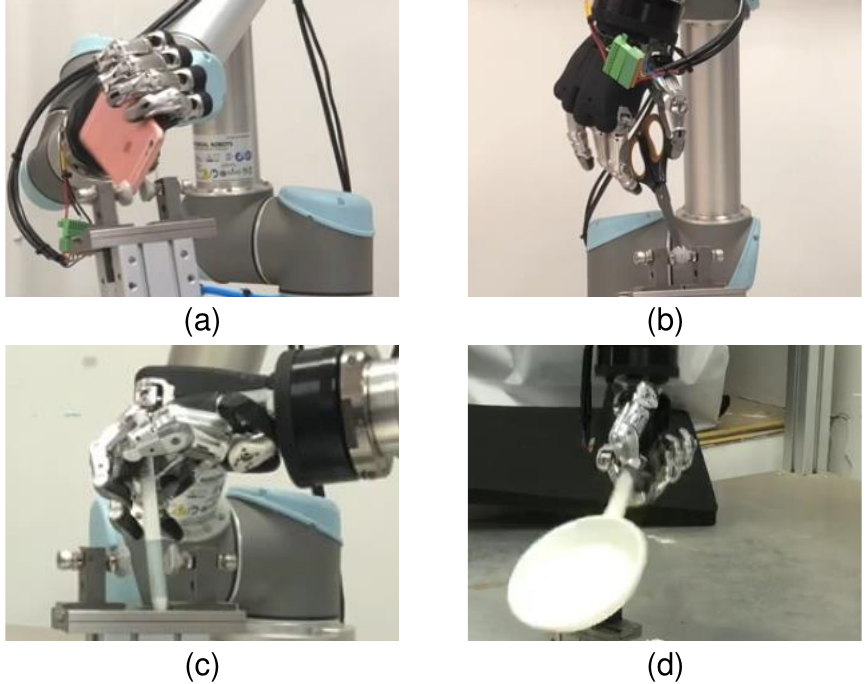}
    \caption{Some failed grasping trails.}
    \label{fig:failure_cases}
\end{figure}
In general, failure cases in the physical experiments can be attributed to the following factors: 1) Reconstruction errors caused by the shape completion module; 2) Unstable grasps produced by our DexFG-Net; 3) Uncertainly in joint position control. Most of the time, the failure grasps are a combination of these factors.

Fig. \ref{fig:failure_cases}(a) reports a failure grasp mainly caused by unstable grasp.  Fig. \ref{fig:failure_cases}(b) shows that our algorithm struggles to generate functional grasps for difficult cases, such as scissors with small handle holes for finger wrapping. Fig. \ref{fig:failure_cases}(c)  reports a failure grasp caused by uncertainly in joint position control. Since the Schunk Hand is an under-actuated anthropomorphic hand, the joint coupling relationship is not strictly linear, which causes uncertainty in sim-to-real transfer. The failure grasp shown in Fig. \ref{fig:failure_cases}(d) is mainly caused by inaccurate object state parameters estimation. Additionally, the robot fails to grasp transparent objects because the depth camera cannot provide accurate prediction for transparent surface.

The robotic setup shown in Fig. \ref{fig:real_robot_platform} indicates  that our approach requires the object to be clamped by a pneumatic gripper in order to perform functional grasps with anthropomorphic hands. However, picking up the tool directly from the table and precisely adjusting its position through in-hand manipulation, akin to human dexterity, still presents a significant challenge.


\section{Conclusion}
In summary, this paper presents a novel framework that can synthesize physically plausible and human-like functional tool-use grasp with minimal demonstrations. The proposed framework achieves this by utilizing fine-grained contact modeling, which allows us to generate grasps for a wide range of kinematically diverse hand models. The use of category-level dense shape correspondence facilitates the acquisition of diffused contact maps for category-level objects, significantly reducing the need for human demonstration. Extensive experiments are conducted to demonstrate that the proposed framework outperforms state-of-the-art methods in terms of quantitative metrics, qualitative visualization and user studies, providing a much closer approximation to human grasping with anthropomorphic hands. Finally, we discuss the failure modes of the proposed framework. Overall, our framework has the potential to advance the field of robotic functional grasping using anthropomorphic hands.

\appendix
\subsection{Dataset Collection\label{appendix:dataset_colleciton}}
We collect 2500+ object models of 37 categories to construct the object dataset. Nearly 1500 of these objects are collected from ShapeNet dataset\cite{chang2015shapenet}, YCB-Video dataset\cite{ycb}, others collected from the online GrabCAD library. Objects in the dataset are classified as follows: 
\begin{itemize}
    \item Containers: Can, Detergent bottle, Lotion bottle, Spray bottle,  Squeeze Bottle,  Water bottle.
    \item Electronic Devices: Camera, Headphones,  Mouse, Phone, Xbox.  
    \item Kitchenware: Bowl, Cup,  Fork, Knife, Mug, Pan, Spatula, Spoon, Teapot,  Wine glasses.
    \item Illumination devices: Flashlight, Lightbulb.
    \item Office supplies: Clamp, Pen, Stapler.
    \item Personal Items: Eyeglasses, Key, Toothbrush,  Toothpaste.
    \item Tools: Hammer,  Pliers, Power drill, Scissors, Screwdriver, Wrench. 
    \item Others: Binoculars, Doorknob.
\end{itemize}

\subsection{Anchor Points of Robot Hands\label{appendix:anchor_points}}
\begin{figure}[h]
    \centering
    \includegraphics[width=1.0\linewidth]{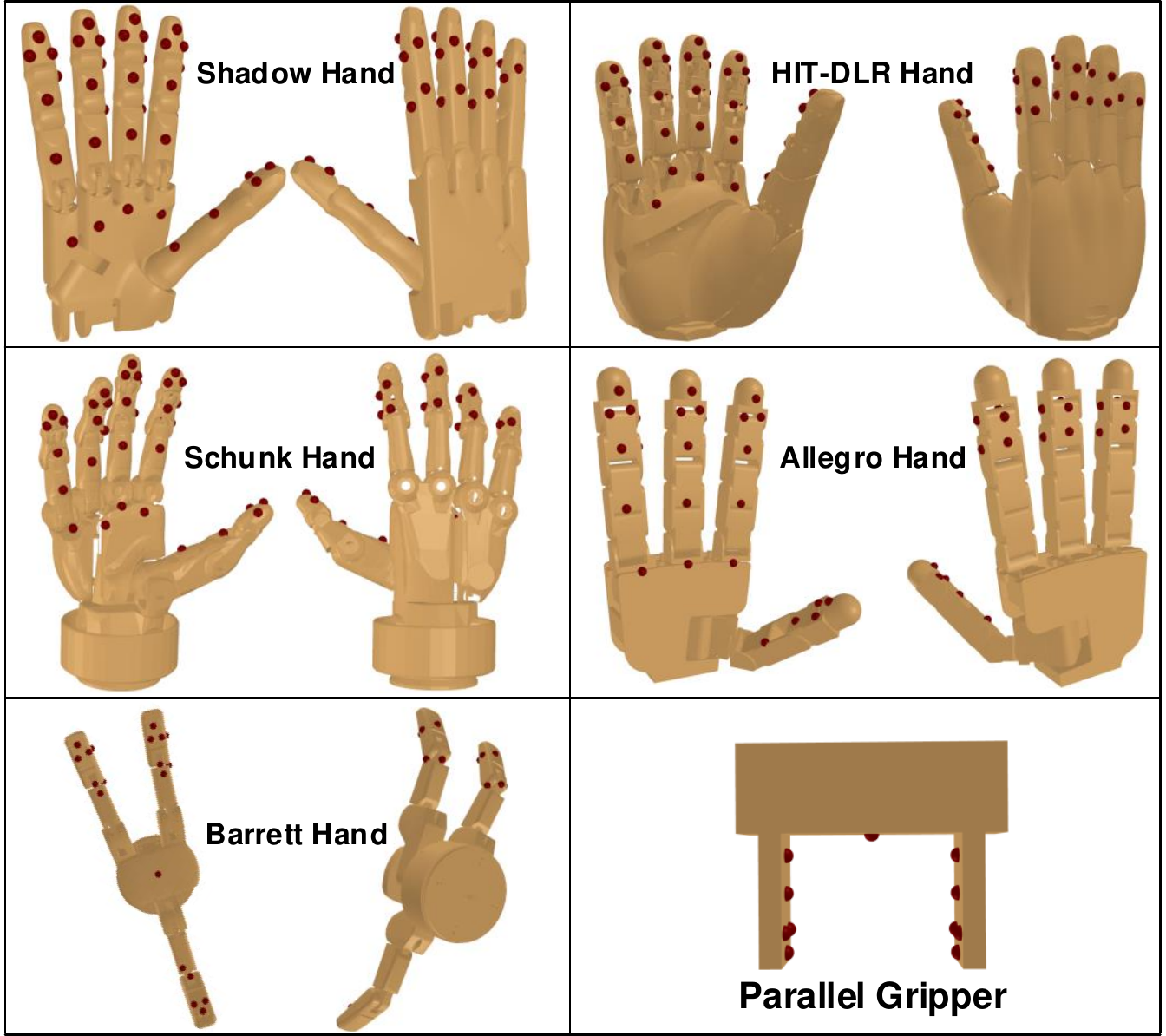}
    \caption{Anchor points distribution of the six robot hand models (in red).}
    \label{fig:robot_anchor_points}
\end{figure}

Anchor points on robot hand models are manually annotated according to the position of anchor points on human hand. Fig. \ref{fig:robot_anchor_points} shows the distribution of anchor points on the six hand models used in our functional grasp synthesis algorithm.



\ifCLASSOPTIONcaptionsoff
  \newpage
\fi


\bibliographystyle{IEEEtran}
\bibliography{reference}

\end{document}